\documentclass[acmsmall]{acmart}

\usepackage{hyperref}
\usepackage{url}

\usepackage[utf8]{inputenc} 
\usepackage[T1]{fontenc}    
\usepackage{booktabs}       
\usepackage{amsfonts}       
\usepackage{nicefrac}       
\usepackage{microtype}      
\usepackage{xcolor}         
\usepackage[shortlabels]{enumitem}
\usepackage{lipsum}
\usepackage{latexsym}

\usepackage{siunitx}

\usepackage{amssymb}
\usepackage{amsmath}
\usepackage{amsthm}
\usepackage{float}
\usepackage{graphicx}
\usepackage{color}
\usepackage{multirow}
\usepackage{caption}

\usepackage{longtable}

\usepackage{colortbl}
\usepackage{enumitem}
\usepackage{array}
\usepackage{xcolor}

\theoremstyle{definition}
\newtheorem{definition}{Definition}[section]

\AtBeginDocument{%
  }

\setcopyright{acmlicensed}
\copyrightyear{2025}
\acmYear{2025}
\acmDOI{XXXXXXX.XXXXXXX}

\acmJournal{JACM}

\usepackage{amsmath,amsfonts,bm}

\def\eqref#1{equation~\ref{#1}}

\def\1{\bm{1}}

\DeclareMathAlphabet{\mathsfit}{\encodingdefault}{\sfdefault}{m}{sl}
\SetMathAlphabet{\mathsfit}{bold}{\encodingdefault}{\sfdefault}{bx}{n}

\begin{document}


\title{Spectraformer: A Unified Random Feature Framework for Transformer}



\author{Duke Nguyen}
\email{research@itsduke.me}
\orcid{0009-0006-4445-1534}
\author{Du Yin}
\orcid{0000-0002-2345-0683}
\email{du.yin@unsw.edu.au}
\author{Aditya Joshi}
\orcid{0000-0003-2200-9703}
\email{aditya.joshi@unsw.edu.au}
\author{Flora Salim}
\orcid{0000-0002-1237-1664}
\email{flora.salim@unsw.edu.au}
\affiliation{%
  \institution{University of New South Wales} 
  \city{Sydney}
  \state{New South Wales}
  \country{Australia}
}


\begin{abstract}
Linearization of attention using various kernel approximation and kernel learning techniques has shown promise. Past methods used a subset of combinations of component functions and weight matrices within the random feature paradigm. We identify the need for a systematic comparison of different combinations of weight matrices and component functions for attention learning in Transformer. Hence, we introduce \textit{Spectraformer}, a unified framework for approximating and learning the kernel function in the attention mechanism of the Transformer. Our empirical results demonstrate, for the first time, that a random feature-based approach can achieve performance comparable to top-performing sparse and low-rank methods on the challenging Long Range Arena benchmark. Thus, we establish a new state-of-the-art for random feature-based efficient Transformers. The framework also produces many variants that offer different advantages in accuracy, training time, and memory consumption. Our code is available at: \url{https://github.com/cruiseresearchgroup/spectraformer}.

\end{abstract}




\maketitle

\section{Introduction}\label{sec:intro}
Transformer \citep{vaswani2017attention} has revolutionized the landscape of natural language processing (NLP) and forms the basis of almost all state-of-the-art language models. Its influence has reached beyond NLP into computer vision~\citep{han2022survey, yu2024rethinking, 10.1007/978-3-031-72973-7_25, 10844040}, speech processing~\citep{lin2022survey}, and other fields. Compared to its precursors like the LSTMs, Transformer leverages parallelism since it is fully based on the attention mechanism. Therefore, the performance of Transformer is tied to the effective use of attention. Attention in the Original Transformer (referred to as `\textbf{OT}' hereafter) uses the softmax function which is quadratic in time complexity. 

To mitigate the quadratic complexity of the standard attention mechanism, a diverse range of efficient Transformer architectures have been developed~\citep{tay_survey}. These include memory/ downsampling methods (Big Bird~\citep{zaheer2020big}, Nystromformer~\citep{nystromformer}, and Skyformer~\citep{chen2021skyformer}), learnable pattern methods (Reformer~\citep{kitaev2020reformer}), Informer~\citep{informer}, Linformer~\citep{wang2020linformer}. Our work, \textit{Spectraformer}, focuses on another significant category: the linearization of attention through a unified framework of random features. We approach this linearization by rethinking the role of the softmax. The softmax acts as a \textbf{compatibility function}, capturing the similarity between pairs of tokens (see Equations \ref{eq:attention2} and \ref{eq:compatibility_softmax}). Extensive experiments with other compatibility functions show that the softmax is not the only possibility, and that the `best' \textbf{compatibility function} depends on the task \citep{tsai2019transformer}.

Replacing the softmax formulation with alternatives may allow us to reduce the time complexity of attention computation. In this paper, we focus on the subset of alternatives that employs kernel functions. Kernel functions have been historically used in machine learning algorithms to simplify the estimation of parameters~\citep{hofmann2008kernel}. \citet{tsai2019transformer} show that attention can be formulated using kernels and then linearized \citep{katharopoulos20a}, given the feature map of the respective attention kernel. Random feature-based algorithms \citep{liu2021random} are a family of algorithms, inspired by spectral analysis, which provides the feature map associated with a kernel. A random feature consists of three components: component function, weight matrix, and component weight (see Section \ref{sec:method}). Due to their robustness and flexibility, random features give rise to `\textit{random feature Transformers}', the family of Transformers with linearized attention via random features (hereby abbreviated as RF), first with the work by \citet{Choromanski2020RethinkingAW}. This is seen to lead to three strands of research: (A) \textbf{Optimization of weight matrices} either (i) by making the matrices have lower time or space complexity, or (ii) by incorporating beneficial properties like orthogonality to reduce variance \citep{Yu2016OrthogonalRF, Choromanski2020RethinkingAW, choromanski2022hybrid, reid2023simplex}; (B) \textbf{Enhancement of component functions} by either: (i) finding a function with a tighter variance or approximation bound, or (ii) engineering in order to output with numerical stability and be bounded \citep{Choromanski2020RethinkingAW, crt2022, likhosherstov2023dense}; (C) Parameterization of weight matrices to perform kernel learning instead of kernel approximation \citep{chowdhury2022learning}. (A) and (B) seek to improve approximation quality via approximation error and variance reduction. (C) dispenses with the approximation of the attention kernel (typically the softmax), and makes the kernel itself learnable, to attain a better performance. However, the three have been explored separately, creating a situation where there are different ways to construct the random features. Additionally, past works only explore certain combinations of component functions and weight matrices, creating disjoint overlaps and gaps in the literature. Finally, the weight matrices are often those explicitly designed in the context of kernelized attention in the Transformer \citep{Yu2016OrthogonalRF, Choromanski2020RethinkingAW, choromanski2022hybrid, reid2023simplex}, ignoring those which are popular in the kernel method literature (e.g., \citet{liu2021random}). These limitations highlight the need for a unified framework to systematically compare combinations and identify optimal configurations.



\begin{table*}[ht!]
    \caption{Experimental results of \textit{Spectraformer} variants against previous SOTA efficient Transformers on the LRA benchmark. All models were run for five seeds using 128 random features. We report mean accuracy (on the test set), mean time (training time (hour)), and mean memory (peak memory consumption (GB)). $L$: ListOps, $T$: Text, $R$: Retrieval, $I$: Image, $P$: Pathfinder, and $\mu$ for the average across tasks. Entries are sorted in descending order by $\mu$. Best models per category are boldened.}
    \label{table:first_table}
    \centering
    \resizebox{\linewidth}{!}{
    \begin{tabular}{lccccccccccccccccccc}
    \toprule
    & \multicolumn{6}{c}{Accuracy (\%) $\uparrow$} & \multicolumn{6}{c}{Time (hour) $\downarrow$} & \multicolumn{6}{c}{Memory (GB) $\downarrow$} \\ 
    \cmidrule(lr){2-7} \cmidrule(lr){8-13} \cmidrule(lr){14-19}
    & L & T & R & I & P & $\mu$ & L & T & R & I & P & $\mu$ & L & T & R & I & P & $\mu$ \\
    \midrule
\textbf{Nystromformer}           & 38.69 (0.59)          & 61.57 (0.38)          & 80.57 (0.30)          & 39.49 (0.89)          & \textbf{69.96 (1.27)} & \textbf{58.06} & 0.56          & 0.90          & 1.00          & 2.19          & 1.19          & 1.17          & 1.10          & 1.87          & 1.84          & 5.55          & 2.79          & 2.63          \\
Big Bird                         & 38.82 (0.52)          & 61.04 (0.15)          & 80.53 (0.32)          & 37.59 (1.04)          & 68.28 (6.74)          & 57.25          & 1.60          & 3.17          & 3.24          & 5.72          & 2.89          & 3.32          & 2.24          & 4.57          & 3.80          & 8.42          & 4.22          & 4.65          \\
OPRF-FastFood\textsubscript{L}   & 37.73 (0.28)          & 64.32 (0.61)          & 78.65 (1.94)          & 38.41 (0.66)          & 66.76 (0.39)          & 57.17          & 1.07          & 2.07          & 2.11          & 4.02          & 2.05          & 2.26          & 0.84          & 1.68          & 1.64          & 3.26          & 1.68          & 1.82          \\
Reformer                         & 35.50 (4.09)          & 61.28 (0.45)          & 78.31 (0.29)          & \textbf{43.62 (0.81)} & 66.28 (2.35)          & 57.00          & 0.65          & 1.29          & 1.29          & 2.51          & 1.26          & 1.40          & 1.61          & 3.20          & 2.95          & 6.38          & 3.20          & 3.47          \\
Skyformer                        & \textbf{38.96 (0.63)} & 60.67 (0.45)          & \textbf{81.90 (0.37)} & 32.93 (0.39)          & 69.81 (1.38)          & 56.85          & 0.78          & 1.35          & 1.49          & 3.09          & 1.63          & 1.67          & 1.67          & 3.01          & 2.92          & 7.90          & 3.96          & 3.89          \\
OT                               & 38.49 (0.80)          & 59.72 (1.07)          & 80.86 (0.20)          & 37.48 (0.83)          & 67.56 (6.12)          & 56.82          & 1.96          & 7.27          & 7.07          & 4.30          & 2.20          & 4.56          & 4.37          & 16.72         & 8.74          & 9.42          & 4.72          & 8.79          \\
OPRF-SGQ                         & 37.08 (0.56)          & 61.09 (0.77)          & 79.39 (0.94)          & 36.68 (0.28)          & 67.53 (2.79)          & 56.35          & 0.68          & 1.27          & 1.25          & 2.46          & 1.29          & 1.39          & 1.33          & 2.64          & 2.50          & 5.25          & 2.64          & 2.87          \\
PosRF-MM                         & 37.13 (0.27)          & 62.88 (1.17)          & 80.58 (0.44)          & 33.82 (0.51)          & 67.11 (0.45)          & 56.30          & 0.56          & 1.05          & 1.06          & 2.02          & 1.05          & 1.15          & 1.14          & 2.26          & 2.05          & 4.49          & 2.26          & 2.44          \\
OPRF-OR                          & 38.05 (0.78)          & 60.18 (0.49)          & 81.19 (0.23)          & 33.70 (0.57)          & 67.24 (0.88)          & 56.07          & 0.68          & 1.26          & 1.24          & 2.47          & 1.29          & 1.39          & 1.33          & 2.64          & 2.50          & 5.25          & 2.64          & 2.87          \\
SADERF-QMC                       & 37.18 (0.47)          & 60.61 (2.06)          & 80.67 (0.44)          & 34.55 (0.63)          & 67.32 (0.40)          & 56.07          & 0.68          & 1.24          & 1.28          & 2.47          & 1.26          & 1.39          & 1.40          & 2.79          & 2.63          & 5.56          & 2.79          & 3.04          \\
PosRF-QMC                        & 37.05 (0.16)          & 60.93 (0.59)          & 80.67 (0.29)          & 34.03 (0.94)          & 67.25 (0.44)          & 55.99          & 0.55          & 1.05          & 1.05          & 1.99          & 1.05          & 1.14          & 1.14          & 2.26          & 2.05          & 4.49          & 2.26          & 2.44          \\
OPRF-QMC                         & 37.86 (0.40)          & 60.87 (1.85)          & 80.44 (0.20)          & 33.87 (0.69)          & 66.81 (0.68)          & 55.97          & 0.68          & 1.26          & 1.24          & 2.43          & 1.28          & 1.38          & 1.33          & 2.64          & 2.50          & 5.25          & 2.64          & 2.87          \\
SADERF-ORF                       & 37.41 (0.45)          & 59.92 (0.90)          & 81.05 (0.18)          & 33.06 (1.02)          & 67.11 (0.56)          & 55.71          & 0.69          & 1.25          & 1.28          & 2.50          & 1.28          & 1.40          & 1.40          & 2.79          & 2.63          & 5.56          & 2.79          & 3.04          \\
OPRF-MM                          & 38.31 (0.38)          & 60.01 (1.01)          & 81.03 (0.31)          & 33.93 (0.85)          & 65.17 (5.17)          & 55.69          & 0.68          & 1.26          & 1.29          & 2.47          & 1.28          & 1.40          & 1.33          & 2.64          & 2.50          & 5.25          & 2.64          & 2.87          \\
SADERF-MM                        & 37.17 (0.27)          & 60.81 (1.82)          & 81.05 (0.20)          & 34.03 (1.04)          & 65.31 (4.85)          & 55.67          & 0.69          & 1.25          & 1.28          & 2.49          & 1.28          & 1.40          & 1.40          & 2.79          & 2.63          & 5.56          & 2.79          & 3.04          \\
SADERF-SGQ                       & 37.22 (0.28)          & 62.86 (0.96)          & 78.51 (0.68)          & 37.82 (0.59)          & 61.27 (5.38)          & 55.54          & 0.68          & 1.25          & 1.28          & 2.50          & 1.27          & 1.39          & 1.40          & 2.79          & 2.63          & 5.56          & 2.79          & 3.04          \\
SADERF-FastFood\textsubscript{L} & 29.72 (7.27)          & 64.71 (0.45)          & 77.50 (0.96)          & 38.38 (0.82)          & 66.47 (1.47)          & 55.35          & 1.09          & 2.09          & 2.18          & 4.05          & 2.06          & 2.29          & 0.90          & 1.80          & 1.76          & 3.52          & 1.80          & 1.96          \\
PosRF-FastFood\textsubscript{L}  & 29.54 (5.46)          & 64.61 (0.29)          & 77.10 (1.02)          & 38.28 (0.43)          & 66.38 (0.71)          & 55.18          & 1.02          & 2.00          & 2.01          & 3.88          & 1.98          & 2.18          & \textbf{0.77} & \textbf{1.54} & \textbf{1.50} & \textbf{3.01} & \textbf{1.54} & \textbf{1.67} \\
PosRF-ORF                        & 34.50 (6.49)          & 61.10 (1.76)          & 80.53 (0.33)          & 33.72 (1.03)          & 65.52 (4.89)          & 55.07          & 0.56          & 1.05          & 1.06          & 2.02          & 1.06          & 1.15          & 1.14          & 2.26          & 2.05          & 4.49          & 2.26          & 2.44          \\
OPRF-SORF                        & 29.59 (3.77)          & 64.81 (0.22)          & 77.12 (0.78)          & 37.42 (0.47)          & 63.98 (5.38)          & 54.58          & 0.68          & 1.26          & 1.24          & 2.42          & 1.28          & 1.38          & 1.33          & 2.64          & 2.50          & 5.25          & 2.64          & 2.87          \\
SADERF-SORF                      & 34.31 (2.50)          & \textbf{64.82 (0.26)} & 75.24 (1.12)          & 36.76 (1.22)          & 60.98 (4.90)          & 54.42          & 0.68          & 1.25          & 1.28          & 2.47          & 1.26          & 1.39          & 1.40          & 2.79          & 2.63          & 5.56          & 2.79          & 3.04          \\
Linformer                        & 36.94 (0.60)          & 56.99 (1.23)          & 77.86 (0.35)          & 39.09 (0.78)          & 59.24 (5.84)          & 54.02          & \textbf{0.41} & \textbf{0.74} & \textbf{0.77} & \textbf{1.28} & \textbf{0.71} & \textbf{0.78} & 0.86          & 1.70          & 1.61          & 3.39          & 1.71          & 1.85          \\
PosRF-SGQ                        & 28.81 (7.77)          & 62.30 (0.47)          & 78.31 (0.39)          & 37.96 (0.76)          & 59.71 (7.87)          & 53.42          & 0.56          & 1.06          & 1.05          & 2.01          & 1.06          & 1.15          & 1.14          & 2.26          & 2.05          & 4.49          & 2.26          & 2.44          \\
Informer                         & 31.45 (7.20)          & 62.06 (1.20)          & 76.92 (0.14)          & 37.90 (0.30)          & 57.07 (6.57)          & 53.08          & 0.88          & 1.92          & 2.23          & 3.21          & 1.78          & 2.01          & 2.66          & 5.57          & 3.27          & 5.20          & 2.61          & 3.86          \\
PosRF-SORF                       & 21.91 (5.98)          & 62.06 (1.36)          & 66.33 (0.41)          & 28.70 (3.87)          & 52.37 (5.24)          & 46.27          & 0.55          & 1.06          & 1.06          & 2.00          & 1.05          & 1.14          & 1.14          & 2.26          & 2.05          & 4.49          & 2.26          & 2.44          \\ \bottomrule
\end{tabular}
}
\end{table*}

We introduce \textit{Spectraformer}, a unified framework for approximating and learning the kernel function in the attention mechanism of the Transformer through the lens of random features. While other methods (e.g., Nystromformer~\citep{nystromformer}, BigBird~\citep{zaheer2020big}) have explored alternative approaches to improving efficiency, \textit{Spectraformer} provides a systematic way to explore the vast design space of random feature-based attention. It utilizes \textbf{spectral} analysis-inspired random features to construct linearized attention in \textbf{Transformers}. \textit{Spectraformer} allows for the experimentation of various combinations of weight matrices and component functions. Through our benchmarking experiments on the \textbf{Long Range Arena} (LRA)~\citep{tay2021long} and \textbf{18 combinations} of weight matrices and component functions, \textit{Spectraformer} produces novel combinations of component functions and weight matrices that perform competitively against previous efficient Transformers with different computational advantages. In addition, our work demonstrates that a well-configured kernelized attention model can perform on the same level as other classes of efficient Transformers, with our best \textit{Spectraformer} variant proving highly competitive with leading methods like Nystromformer~\citep{nystromformer} and BigBird~\citep{zaheer2020big}. In doing so, it establishes a new state-of-the-art for random feature-based methods with a mean accuracy improvement of 0.35\% over the Original Transformer (OT). Many \textit{Spectraformer} variants also outperform existing efficient Transformers with random feature attention such as SADERF-ORF (a.k.a., FAVOR\# \citep{likhosherstov2023dense}), OPRF-ORF (a.k.a., FAVOR++~\citep{crt2022}), PosRF-ORF (a.k.a., FAVOR+~\citep{Choromanski2020RethinkingAW}). Our contributions are:
\begin{enumerate}
    \item A novel framework, \textit{Spectraformer}, that unifies and generalizes past work on linearized attention using both kernel approximation and kernel learning.
    \item Establishment of a new state-of-the-art for random feature-based efficient Transformers, demonstrating that this model class can be highly competitive with leading sparse and low-rank methods on the Long Range Arena benchmark.
    \item A comprehensive empirical analysis of 18 combinations of component functions and weight matrices, offering practical guidance on selecting configurations based on trade-offs between accuracy, memory, and training time.
    \item A public and extensible implementation designed to facilitate future research by allowing for the easy integration of new kernel-based techniques into Transformers.
\end{enumerate}

\section{Preliminaries}\label{sec:prelim}
We first cover the theoretical foundations of this work.

\subsection{Kernelized Attention}\label{sec:attention}
OT attention~\citep{vaswani2017attention} can be defined using the unified attention model~\citep{galassi} as follows. Given a collection of inputs and a learning objective, \textbf{attention} mechanism learns the attention matrix $\textbf{A}$ which captures the associative information between pairs of tokens in a sequence of length $N$, with each row $\textbf{A}_i$ being the representation of a token $i$ with every other token $j \in [1,N]$. This is done by creating three learnable representations of each token $i$ called \textbf{query} ($\boldsymbol{q}_i$), \textbf{key} ($\boldsymbol{k}_i$), and \textbf{value} ($\boldsymbol{v}_i$). The attention representation $\textbf{A}_i$ is calculated as a weighted sum of value $\boldsymbol{v}_j$ given \textbf{attention weights} $a_j$. The general equation for $\textbf{A}_i$ is given as:

\begin{equation}\label{eq:attention2}
\begin{array}{cccc}
\boldsymbol{A}_i =\sum_{j=1}^{N}\frac{f(\boldsymbol{q}_i, \boldsymbol{k}_j)}{\sum_l f(\boldsymbol{q}_i, \boldsymbol{k}_l)}\boldsymbol{v}_j = \sum_{j=1}^N a_j \boldsymbol{v}_j ; & a = g(e) ; & e = f(\boldsymbol{q}_i, \boldsymbol{k}_j)
\end{array}
\end{equation}

Attention weights are generated using a \textbf{distribution function} $g$ applied on an energy score $e$. The energy score is captured by applying a \textbf{compatibility function} on a query $\boldsymbol{q}_i$ and a set of keys $\textbf{K}$, in order to compute a score of a specific relationship between the pairs. The attention in the OT is called the `scaled dot-product' which defines the compatibility function $f$ and the distribution function $g$ as follows:
\begin{equation}\label{eq:compatibility_softmax}
\begin{array}{cc}
    f(\boldsymbol{q}_i, \textbf{K}) = \frac{\boldsymbol{q}_i^T \textbf{K}}{\sqrt{d_k}} & g(\boldsymbol{x}_j) = \frac{\exp(\boldsymbol{x}_j)}{\sum_l \exp(\boldsymbol{x}_l)}
\end{array}
\end{equation}
We can then have the attention matrix (with $\exp(.)$ being applied element-wise) as:
\begin{equation}\label{eq:softmax_attention}
\begin{array}{cc}
    \boldsymbol{A}_i = \sum_{j=1}^{N}\frac{\exp(\boldsymbol{q}_i \boldsymbol{k}_j^T/\sqrt{d_k})}{\sum_l \exp(\boldsymbol{q}_i \boldsymbol{k}_l^T/\sqrt{d_k})} \boldsymbol{v}_j & A = softmax(\frac{\boldsymbol{QK}^T}{\sqrt{d_k}})\boldsymbol{V}
\end{array}
\end{equation}
with $\boldsymbol{Q}, \boldsymbol{K}, \boldsymbol{V}$ being the matrix of queries, keys, and values respectively with each row being $\boldsymbol{q}_i, \boldsymbol{k}_i, \boldsymbol{v}_i$.

We now discuss the intrinsic connection between kernel and attention. With the compatibility function $f$ as a kernel $f(\boldsymbol{q}_i, \boldsymbol{k}_j) = \mathcal{K}(\boldsymbol{q}_i, \boldsymbol{k}_j)$, Equation \ref{eq:attention2} can be rewritten as:

\begin{equation}\label{eq:attention-row-kernel}
    \boldsymbol{A}_i=\sum_{j=1}^{N}\frac{\mathcal{K}(\boldsymbol{q}_i, \boldsymbol{k}_j)}{\sum_l \mathcal{K}(\boldsymbol{q}_i, \boldsymbol{k}_l)}\boldsymbol{v}_j
\end{equation}

As shown by \citet{tsai2019transformer}, this is indeed the Nadaraya-Watson Kernel Estimator, with \\$\mathbb{E}_{p(\boldsymbol{k}_j|\boldsymbol{q}_i)}[\boldsymbol{v}_j|X=\boldsymbol{k}_j] = \boldsymbol{A}_i$, $l_{i}\left(\boldsymbol{k}_j)=p(\boldsymbol{k}_j|\boldsymbol{q}_i\right)$, $Y_i = \boldsymbol{v}_j$:
\begin{equation}
\begin{array}{ccc}\mathbb{E}_{l_{i}}[Y_{i}|X=\boldsymbol{u}] = \sum_{i=1}^{N}l_{i}(\boldsymbol{u})Y_{i} ; & l_{i}(\boldsymbol{u})=\frac{\mathcal{K}(\frac{\boldsymbol{u}-\boldsymbol{x}_i}{h})}{\sum_{j=1}^n \mathcal{K}(\frac{\boldsymbol{u}-\boldsymbol{x}_j}{h})}\end{array}
\end{equation}
Specifically, the scaled dot-product attention in the OT in Equation \ref{eq:softmax_attention} corresponds to the softmax kernel (see Equation \ref{softmax-kernel}).

\begin{equation}\label{softmax-kernel}
    \mathcal{K}_{softmax}(\boldsymbol{q}_i, \boldsymbol{k}_j) = \exp(\boldsymbol{q}_i^T \boldsymbol{k}_j/\sqrt{d_k})
\end{equation}

\subsection{Linearizing Kernelized Attention via Random Features}\label{sec:linearized_attention}
Since a kernel is the inner product of two vectors in some space with respect to some feature map $\phi$, i.e., $\mathcal{K}(\boldsymbol{x},\boldsymbol{y}) = \phi(\boldsymbol{x}) \phi(\boldsymbol{y})^T$, then we can rewrite Equation \ref{eq:attention-row-kernel} as:

\begin{equation}\label{eq:linearized-attention}
\begin{array}{c}
\boldsymbol{A}_i=\frac{\sum_j \phi(\boldsymbol{q}_i) \phi(\boldsymbol{k}_j)^T \boldsymbol{v}_j}{\sum_l \phi(\boldsymbol{q}_i) \phi(\boldsymbol{k}_l)^T} =\frac{\phi(\boldsymbol{q}_i) \sum_j \phi(\boldsymbol{k}_j) \otimes \boldsymbol{v}_j}{\phi(\boldsymbol{q}_i) \sum_l \phi(\boldsymbol{k}_l)^T}
\end{array}
\end{equation}

If we pre-compute $\sum_j \phi(\boldsymbol{k}_j) \otimes \boldsymbol{v}_j$ and $\sum_l \phi(\boldsymbol{k}_l)^T$, the entire term becomes $O(1)$ and we only need to compute Equation \ref{eq:linearized-attention} $N$ times for each $i$, resulting in an attention matrix calculated in linear time $O(N)$. We now refer to Equation \ref{eq:linearized-attention} as \textbf{linearized attention} \citep{katharopoulos20a}.

All that is left is to retrieve the feature map $\phi$, the process of which is called \textbf{kernel approximation} (see Section \ref{sec:random-features}), which corresponds to the kernels we specify. There are many popular kernels from which we can choose, but typically, we want to approximate the softmax kernel in Equation \ref{softmax-kernel}. However, since the softmax relates to the $\mathcal{K}_{RBF}(\boldsymbol{x},\boldsymbol{y}) = \exp(-\frac{||\boldsymbol{x}-\boldsymbol{y}||^2}{2})$ via the following equation:
\begin{equation}\label{eq:softmax_from_gaus}
    \mathcal{K}_{softmax}(\boldsymbol{x},\boldsymbol{y}) =\exp(\frac{||\boldsymbol{x}||^2}{2})\mathcal{K}_{RBF}(\boldsymbol{x},\boldsymbol{y})\exp(\frac{||\boldsymbol{y}||^2}{2})
\end{equation}
Approximating the RBF is easier than the softmax directly in the method we introduce in Section \ref{sec:random-features}, hence we approximate the softmax indirectly via the RBF in Equation \ref{eq:softmax_from_gaus}.

Oftentimes, however, we might not want to specify a kernel, since kernel choice is indeed a hyperparameter and the softmax is by no means essential or necessary as shown by \citet{tsai2019transformer}. Then kernel learning would be a more suitable option, the detail of which is discussed in Section \ref{sec:kernel_learning_intro}. Notwithstanding, we first introduce the main kernel approximation technique: random features.

\subsection{Random Features for Kernel Approximation} \label{sec:random-features}
Of the kernel approximation family of methods, the most successful and appropriate for linearized attention in Transformer has been the random features approach \citep{Rahimi2007RandomFF}. Estimating kernels using random features relies on a fundamental insight from harmonic analysis called Bochner's theorem \citep{rudin2017fourier} stated as follows:

\textit{A continuous shift-invariant kernel $\mathcal{K}(\boldsymbol{x},\boldsymbol{y}) = \mathcal{K}(\boldsymbol{x}-\boldsymbol{y})$ on $\mathbb{R}^d$ is positive definite if and only if $\mathcal{K}(\boldsymbol{\delta})$ is the Fourier transform of a non-negative measure $p(\boldsymbol{\omega})$. If $\mathcal{K}(\boldsymbol{\delta})$ is properly scaled, that is $\mathcal{K}(0) = 1$ then the measure $p(\boldsymbol{\omega})$ is a proper probability distribution.}

\begin{equation}\label{eq:bochner}
    \mathcal{K}(\boldsymbol{\delta})=\int p(\boldsymbol{\omega})\exp(i\boldsymbol{\omega\delta})\mathrm{d}\boldsymbol{\omega}
\end{equation}
When approximating the integration in this equation via Monte Carlo sampling with $s$ samples, $\boldsymbol{\omega} \sim p(\boldsymbol{\omega})$, we can obtain the feature map:
\begin{equation}\label{eq:random_features}
\begin{array}{ll}
     \mathcal{K}(\boldsymbol{\delta})&=\mathbb{E}[\phi_{\boldsymbol{\omega}}(\boldsymbol{x})^T \phi_{\boldsymbol{\omega}}(\boldsymbol{y})] \approx <\phi_{\boldsymbol{\omega}}(\boldsymbol{x}), \phi_{\boldsymbol{\omega}}(\boldsymbol{y})>\\
     \phi_{\boldsymbol{\omega}}(\boldsymbol{x}) \! &= \! \frac{1}{\sqrt{s}} \big [\exp (-\mbox{i}{\boldsymbol{\omega} }^{T }_1 \boldsymbol{x}), \ldots,\exp (-\mbox{i}{\boldsymbol{\omega} }^{T }_s \boldsymbol{x})]^T
\end{array}
\end{equation}

\subsection{Random Features for Kernel Learning}\label{sec:kernel_learning_intro}
Previous random feature techniques are robust in approximating kernels for specific learning problems. However, kernels may be chosen using heuristics and by convention from a popular subset. The kernel is very much a hyperparameter. This could pose a problem since there is no free lunch in learning as much as in kernel choice \citep{Scholkopf2005LearningWK}. \citet{tsai2019transformer} have shown experimental results that this is also true in the context of attention in Transformer. Hence, instead of picking a kernel for the attention, we could learn the kernel via \textbf{kernel learning} an established kernel methods technique. Kernel learning has shown to be effective in general \citep{tompkins2019bbq, wilson2013gaussian, oliva2015bayesian} and in the case of attention via Gaussian Mixture Model (GMM), learnable weight matrices (FastFood), deep generative model (DGM) \citep{chowdhury2022learning}. Kernel learning is achieved by making the weight matrix $\boldsymbol{W}$ learnable where $\boldsymbol{\omega}_i$ is the output of a function parameterizing $p(\boldsymbol{\omega})$.

The kernelized attention via random features discussed here is a popular approach, however it is by no means the only one. A more detailed discussion is offered in related work in Section \ref{appendix:related_work}. Specifically, Section \ref{appendix:kernelized_attention} provides alternative kernelized attention formulations and Section \ref{appendix:kernel_approx} provides alternative kernel approximations to random features.

\begin{figure*}[ht]
    \centering
    \includegraphics[width=\textwidth]{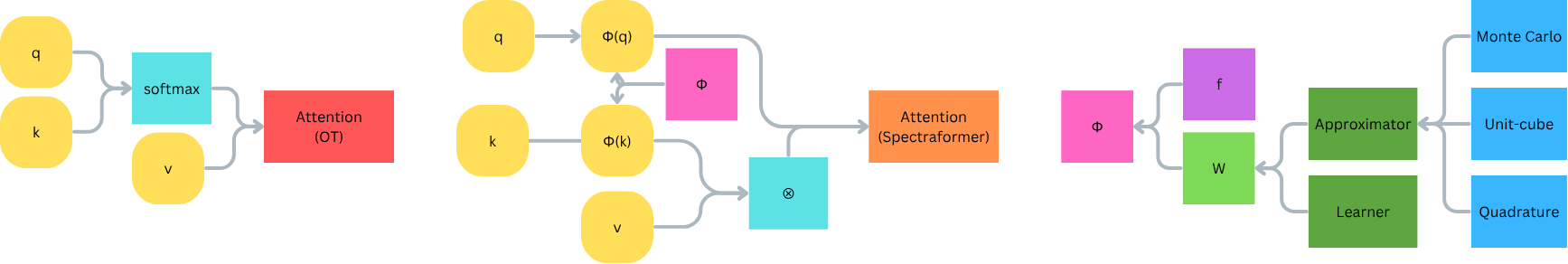}
    \caption{Spectraformer framework based on Equation \ref{eq:rf_general_more}}
    \label{fig:spectraformer_framework}
\end{figure*}

\section{Spectraformer}\label{sec:method}
\textit{Spectraformer} is a unified random feature framework that allows for the combination of any weight matrix with any component function to be used as kernelized attention in Transformer. The general equation for random feature is shown in Equation \ref{eq:rf_general_more}, generalized from Equation \ref{eq:random_features}, based on \citet{liu2021random, Choromanski2020RethinkingAW}.

\begin{equation}
\begin{array}{rl}
\phi_{\boldsymbol{\omega}} (\boldsymbol{x}) = &\frac{1}{\sqrt{s}} \big [a_1 f_1(\boldsymbol{\omega}_1, \boldsymbol{x}), ..., a_s f_1(\boldsymbol{\omega}_s, \boldsymbol{x}), ..., 
\\&a_1 f_l(\boldsymbol{\omega}_1, \boldsymbol{x}), ..., a_s f_l(\boldsymbol{\omega}_s, \boldsymbol{x})]^{\!\top }\\
\boldsymbol{W} = & [\boldsymbol{\omega}_1,...,\boldsymbol{\omega}_{s}]^{T}\in\mathbb{R}^{s*d}
\end{array}
\label{eq:rf_general_more}
\end{equation}

\textit{Spectraformer} is shown in Figure \ref{fig:spectraformer_framework}. The attention of the OT (on the left-hand side) is replaced by the attention of the \textit{Spectraformer} which pre-computes the Hadamard (element-wise) product of values $\boldsymbol{v}$ and transformed keys $\phi(\boldsymbol{k})$ scaled by $\phi(\boldsymbol{k})$, the product of which is multiplied with the transformed queries $\phi(\boldsymbol{q})$. The linear map $\phi$ is composed from weight matrices and component functions. The linear map has $s$ components with $a$ being the component weight (see Equation \ref{eq:rf_general_more}). The component weight $a$ can either be fixed (typically at $a=1$) or learnable as proposed by various kernel works \citep{liu2021random}. We set the component weight a=1 to ensure a consistent and fair comparison with prior random feature Transformers (e.g., \citep{crt2022}).

The figure and the equation together highlight the three strands of research as mentioned previously in Section \ref{sec:intro}: (A) Weight matrix approximator: constructing $\boldsymbol{W}$ more effectively; (B) Weight matrix learner: parameterizing $\boldsymbol{W}$ instead of sampling; (C) Component function: constructing $f_j$ more effectively. 

\subsection{Weight Matrices}\label{sec:method_weight}
In \textit{Spectraformer}, the weight matrix $\boldsymbol{W}$ is either an approximator (from the families of \textit{Monte Carlo sampling} (e.g., ORF and SORF~\citep{Yu2016OrthogonalRF}), \textit{Unit-cube sampling} (e.g., QMC~\citep{avron2016quasi}, MM~\citep{shen2017random}), \textit{Quadrature} (e.g., SGQ~\citep{dao2017gaussian}). or a \textit{learner} (e.g., FastFood\textsubscript{L}~\citep{chowdhury2022learning}). 

We term the matrix $\boldsymbol{W}$ as weight matrix instead of random matrix since although $\boldsymbol{W}$ acts like a random matrix ($\boldsymbol{\omega}_i \sim p(.)$, see Equation \ref{eq:random_features}), it is not guaranteed to be one. In the case of weight matrix approximator, they approximate a random matrix and in the case of weight matrix learner, they parameterize a weight matrix, thereby implicitly learning a distribution $p(.)$ associated with such $\boldsymbol{W}$. The weight matrix is given in Definition \ref{definition:weight_matrix}.

\begin{definition}\label{definition:weight_matrix}
Provided that a matrix can substitute for a random matrix in Equation \ref{eq:rf_general_more}, and the solution for this equation given such matrix and $f$ as TrigRF being the solution for Equation \ref{eq:bochner}, i.e., as an unbiased estimator of $\mathcal{K}$, and TrigRF being $f_1 = f_l = \exp(-i\boldsymbol{\omega}^T \boldsymbol{x})$ (see Equation \ref{eq:random_features}) then such a matrix is a weight matrix.
\end{definition}
It follows from Definition \ref{definition:weight_matrix} that for any weight matrix $\boldsymbol{W}$, $f = TrigRF$ is a valid component function. \textit{Spectraformer} \textbf{enables weight matrix approximation} in terms of three families based on how they solve the intractable integral in Equation \ref{eq:bochner}: 

\begin{itemize}
    \item \textbf{Monte Carlo sampling}-based weight matrix involves approximating the integral of Equation \ref{eq:bochner} using Monte Carlo sampling with the solution given by Equation \ref{eq:random_features} \citep{Rahimi2007RandomFF} with $\boldsymbol{W}$ being the `Base' random matrix of $p(.)$, $\boldsymbol{\omega}_i \sim p(.)$. There are additional methods which improve over Base: structural and geometric methods. Structural methods decompose the Base random matrices $P$ into smaller matrices which reduce time and/ or space complexity. Structural matrices includes FastFood\textsubscript{F}~\citep{chowdhury2022learning}, SCRF~\citep{feng2015random} and P-Model~\citep{choromanski2016recycling}. Geometric methods enforce certain geometrical couplings along some dimensions in the construction of random matrices to reduce the approximation error. Geometric methods include: ROM~\citep{choromanski2017unreasonable} (ORF~\citep{Yu2016OrthogonalRF}, SORF~\citep{Yu2016OrthogonalRF}) and SimRF~\citep{reid2023simplex}. In addition to these methods, Monte Carlo has also been adapted as Random Fourier Signature Features (RFSF)~\citep{doi:10.1137/23M1620478} for unbiased estimation of the signature kernel, designed for sequential data. RFSF also has more scalable variants, such as RFSF-DP (Diagonally Projected) and RFSF-TRP (Tensor Random Projection), that use additional dimensionality reduction.

    \item \textbf{Unit-cube sampling}-based weight matrix transforms Equation \ref{eq:bochner} to an integral on the unit cube $[0, 1]^d$, then performs the approximation with uniform and independent points, thus reducing variance. The first approach is QMC~\citep{avron2016quasi}, where $\boldsymbol{W}: \boldsymbol{\omega}_i = \Phi^{-1}(\boldsymbol{t}_i)$, $\Phi$ being the cumulative distribution function (CDF) associated with $p(.)$ and $\boldsymbol{t}_i$ being a low discrepancy sequence. $\Phi^{-1}(\boldsymbol{t}_i) \sim p(.)$ whilst having a lower variance than direct sampling from $p(.)$. MM~\citep{shen2017random,liu2021random} improves over QMC by replacing $\Phi^{-1}$ with a moment matching scheme $\widetilde{\Phi }^{-1}$ on $\Phi^{-1}$.  

    \item \textbf{Quadrature}-based weight matrix uses quadrature rules with non-uniform deterministic weights. The main approach explored is SGQ~\citep{dao2017gaussian} which uses one-dimensional Gaussian quadrature rule by assuming $\mathcal{K}$ factorizes with respect to the dimensions. Smolyak rule is further implemented to alleviate the curse of dimensionality. 

\end{itemize}

\textit{Spectraformer} also \textbf{allows to learn weight matrices} by learning $p(.)$ via parameterizing $\boldsymbol{W}$. \citet{yang2015carte} introduce several of these parametrization schemes, which are then adapted into the attention setting by \citet{chowdhury2022learning}. This comes from a long line of work in kernel learning. The approach to the dual objectives separates the kernel learning methods into two-stage and one-stage methods \citep{liu2021random}. Two-stage methods (e.g., \citep{yu2015compact, wilson2013gaussian, bullins2018notsorandom}) solves the dual objective separately: $\boldsymbol{W}$ is typically first learned via a kernel alignment scheme~\citep{wang2015overview}, then we solve for $\boldsymbol{v}$. One-stage methods (e.g., \citep{chowdhury2022learning, tompkins2019bbq}), on the other hand, solves the dual objective in parallel: $p(.)$ corresponding to $\boldsymbol{W}$ is parameterized then $\boldsymbol{v}$ is solved typically. It is not plausible to apply kernel alignment~\citep{wang2015overview} on Transformer architectures. Therefore, only one-stage methods can be considered in the context of weight matrix learner. Two kernel learning methods, GMM and FastFood\textsubscript{L}, have been experimented in the non-Transformer setting~\citep{yang2015carte}, then adapted into the Transformer setting~\citep{chowdhury2022learning}. DGM (Deep Generative Model) has also been effective in modeling distributions~\citep{Kingma2013AutoEncodingVB}. It too has been adapted by \citet{chowdhury2022learning} for the Transformer setting. \citet{chowdhury2022learning} show the superior performance of FastFood\textsubscript{L} over other kernel learning-based techniques, hence we have only experimented with FastFood\textsubscript{L} among kernel learning techniques.

\begin{figure*}[ht]
    \centering
    \includegraphics[width=\textwidth]{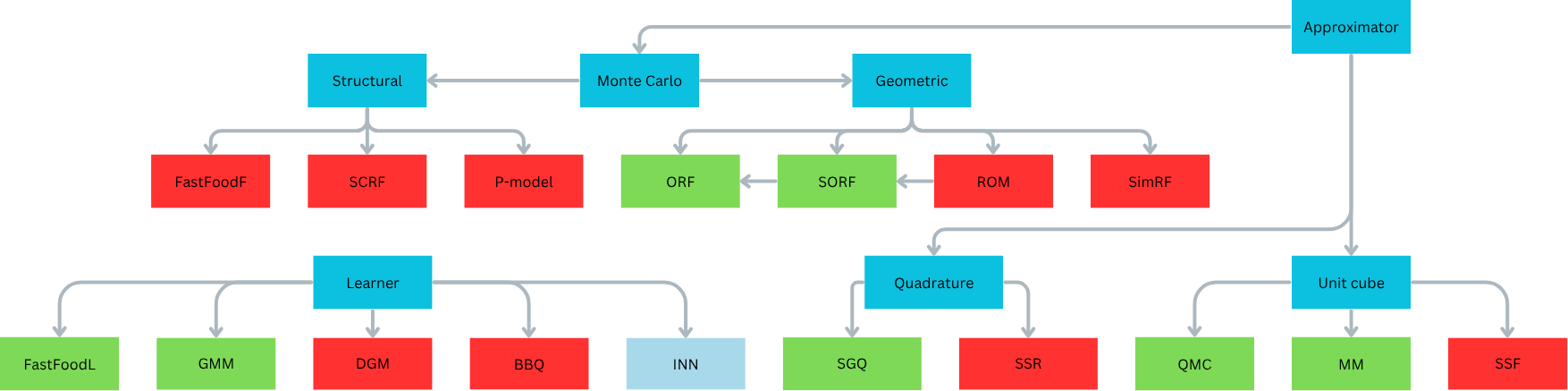}
    \caption{\textit{Spectraformer} weight matrices. Dark blue boxes are technique groupings. Arrow denotes which technique belongs to which grouping. Red boxes are techniques which have been explored in the literature but not in this work. Green boxes are techniques which have been explored in the literature and this work. Light blue boxes are techniques not studied empirically previously.} 
    \label{fig:weight_matrix}
\end{figure*}

\citet{liu2021random} show that in various kernel benchmark datasets (among combination with TrigRF), QMC and ORF have consistently high performance. Due to the significant number of weight matrices in random features literature and page limit, we decide to cover only the most prominent ones. Methods which we will not cover in detail includes: SCRF~\citep{feng2015random}, P-Model~\citep{choromanski2016recycling}, ROM~\citep{choromanski2017unreasonable}, SimRF~\citep{choromanski2017unreasonable}, SSF~\citep{lyu2017spherical}, SSR~\citep{munkhoeva2018quadrature}, GMM~\citep{wilson2013gaussian,oliva2015bayesian}, DGM~\citep{Kingma2013AutoEncodingVB}. We still mention them and include them in the figures where appropriate to motivate future work. Figure \ref{fig:weight_matrix} shows all weight matrices that are implemented and discussed. We now explore these weight matrices in the order specified above.

\subsubsection{Base}~\citep{Rahimi2007RandomFF} This is the direct sampling from a distribution $\boldsymbol{P}_{ij} \sim p$,  corresponding to the kernel $\mathcal{K}$ without any additional adjustment, and in approximating the RBF, $P = G$ where $G$ is the normal distribution.
\begin{equation}\label{eq:gaussian_rff}
    W_{Base} = \frac{1}{\sigma}\boldsymbol{P}
\end{equation}

\subsubsection{FastFood\textsubscript{F}}
FastFood\textsubscript{F}~\citep{le2013fastfood} makes use of Hadamard and diagonal matrices to speed up Gaussian matrices $G$ constructions for RFF in $O(n\log{d})$ (100 times faster) with $O(n)$ space (1000 times less space) given $n \geq d$ against Base. However, FastFood\textsubscript{F} increases variance and approximation error and decreases the concentration bound.
\begin{equation}\label{eq:fastfood}
    W_{FastFood_L} = \frac{1}{\sigma}\boldsymbol{S H G \Pi H B}
\end{equation}
where $\boldsymbol{H}$ is the Walsh-Hadamard matrix, $\boldsymbol{\Pi} \in \{0,1\}^{d*d}$ is a permutation matrix, and $\boldsymbol{S,G,B}$ are diagonal random matrices, with the diagonal entries being $\{+-1\}$ entries on $\boldsymbol{B}$, random Gaussian entries on $\boldsymbol{G}$, and a random scaling matrix on $\boldsymbol{S}$, $\boldsymbol{S}_{ii} = s_i ||\boldsymbol{G}||^{-1/2}_{Frob}$, $s_i \sim (2\pi)^{\frac{-d}{2}} \boldsymbol{A}^{-1}_{d-1} r^{d-1} e^{-\frac{r^2}{2}}$.

\subsubsection{ORF} 
Orthogonal RF~\citep{Yu2016OrthogonalRF} decreases the approximation error significantly compared to Base. This is done via replacing $G$ with a properly scaled random orthogonal matrix. However, generating orthogonal matrices become costly quickly as the number of dimensions increases.
\begin{equation}
    \boldsymbol{W}_{ORF} = \frac{1}{\sigma}\boldsymbol{SQ}
\end{equation}
where $\boldsymbol{Q}$ is a uniformly distributed random orthogonal matrix (on the Stiefel manifold) obtained from the QR decomposition of $\boldsymbol{G}$, the set of rows of $\boldsymbol{Q}$ forming a basis in $\mathbb{R}^d$, and $\boldsymbol{S}$ is a diagonal matrix with entries sampled i.i.d from the $\chi$-distribution with $d$ degrees of freedom, thus making the rows of $\boldsymbol{SQ}$ and $\boldsymbol{G}$ identically distributed.

\subsubsection{SORF}
Structured ORF~\citep{Yu2016OrthogonalRF} decreases time and space complexity of ORF (from $O(d^2)$ to $O(d\log{d})$ with almost no extra memory cost) by imposing structure on the orthogonal matrices, inspired by structural methods. SORF is unbiased with large $d$. We replace $\boldsymbol{S}$ with $\sqrt{d}$ and $\boldsymbol{Q}$ with structured matrix $\boldsymbol{HD}_1 \boldsymbol{HD}_2 \boldsymbol{HD}_3$
\begin{equation}
    \boldsymbol{W}_{SORF} = \frac{\sqrt{d}}{\sigma} \boldsymbol{HD}_1 \boldsymbol{HD}_2 \boldsymbol{HD}_3
\end{equation}
where $\boldsymbol{D}_i \in \mathbb{R}^{d*d}, i = 1,2,3$ is diagonal 'sign-flipping` matrices with diagonal entries sampled from the Rademacher distribution, $\boldsymbol{H}$ is the normalized Walsh-Hadamard matrix.

\subsubsection{QMC}
Quasi-Monte Carlo~\citep{avron2016quasi} evaluates on a low discrepancy sequence (e.g., Halton, Sobol' Faure, and Niederreiter) of points instead of random points in Monte Carlo. Although the approximation error is only reduced minimally, QMC has been shown to perform better than MC in high dimensions and does not have undesirable clustering effect. To calculate QMC, we first assume that $p(\boldsymbol{x}) = \prod^d_{j=1}p_j(\boldsymbol{x}_j)$ factorizes with respect to the dimensions with $p_j(.)$ being a univariate density function. Then we define:
\begin{equation}
\Phi ^{-1}(\boldsymbol{t})=\left(\Phi _{1}^{-1}\left(\boldsymbol t_{1}\right), \ldots, \Phi _{d}^{-1}\left(\boldsymbol t_{d}\right)\right) \in \mathbb {R}^{d}
\end{equation}
where $\Phi_j$ being the cumulative distribution function of $p_j$, $\boldsymbol{t}_1, \boldsymbol{t}_2,...,\boldsymbol{t}_s \in [0,1]^d$ being a low discrepancy sequence, and $\boldsymbol{\omega}_i = \Phi^{-1}(\boldsymbol{t}_i)$. We can thus transform the integral on $\mathbb{R}^d$ in Equation \ref{eq:bochner} to an integral on the unit cube $[0,1]^d$ as
\begin{equation*}\label{eq:unit_cube}
\mathcal{K}(\boldsymbol{x} - \boldsymbol{x}^{\prime }) = \int _{[0,1]^{d}} \exp \big (\mbox{i} (\boldsymbol{x} - \boldsymbol{x}^{\prime })^{\!\top } \Phi ^{-1}(\boldsymbol{t}) \big) \mbox{d} \boldsymbol t
\end{equation*}
The weight matrix then can be defined as:
\begin{equation*}
\boldsymbol W_{\text{QMC}} = [\Phi ^{-1}(\boldsymbol{t}_1), \Phi ^{-1}(\boldsymbol{t}_2), \ldots, \Phi ^{-1}(\boldsymbol{t}_s)]^{\!\top } \in \mathbb {R}^{s \times d}\;.
\end{equation*}
QMC can be further improved by a sub-grouped based rank-one lattice construction which improved complexity \citep{lyu2020subgroupbased}.

\subsubsection{MM} 
Moment Matching~\citep{shen2017random, liu2021random} improves over QMC by removing the undesirable clustering effect and having the same approximation error with less features. This is done by replacing $\Phi ^{-1}$ with a moment matching scheme $\widetilde{\Phi }^{-1}$:
\begin{equation*}
\boldsymbol W_{\text{MM}} = [\widetilde{\Phi }^{-1}(\boldsymbol{t}_1), \widetilde{\Phi }^{-1}(\boldsymbol{t}_2), \ldots, \widetilde{\Phi }^{-1}(\boldsymbol{t}_s)]^{\!\top } \in \mathbb {R}^{s \times d}
\end{equation*}
where $\widetilde{\Phi }^{-1}(\boldsymbol{t}_i) = \tilde{\boldsymbol A}^{-1} (\Phi ^{-1}(\boldsymbol{t}_i) - \tilde{\boldsymbol \mu })$ can be constructed using moment matching with sample mean $\tilde{\boldsymbol \mu } = \frac{1}{s}\sum _{i=1}^{s} \Phi ^{-1}(\boldsymbol{t}_i)$ and the square root of the sample covariance matrix $\tilde{\boldsymbol A}$ satisfying $\tilde{\boldsymbol A} \tilde{\boldsymbol A}^{\!\top } = \text{Cov}(\Phi ^{-1}(\boldsymbol{t}_i) - \tilde{\boldsymbol \mu })$. 


\subsubsection{SGQ}
Sparse Grid Quadrature~\citep{dao2017gaussian} evolves from GQ. GQ (Gaussian Quadrature)~\citep{dao2017gaussian} assumes that the kernel $k$ factorizes with respect to the dimensions and thus can be approximated by a one-dimensional Gaussian quadrature rule. However the total number of points $s$ scale exponentially with the dimensions. However GQ suffers form the curse of dimensionality. This is alleviated using Smolyak rule resulting in SGQ. Assuming the third-degree SGQ using symmetric univariate quadrature points $\left\lbrace -\hat{p}_{1}, 0, \hat{p}_{1}\right\rbrace$ with weights $(\hat{a}_1,\hat{a}_0,\hat{a}_1)$, then we have:
\begin{equation*}
\boldsymbol W_{\text{SGQ}} \!=\! [ \boldsymbol 0_d, \hat{p}_{1} \boldsymbol{e}_{1}, \ldots, \hat{p}_{1} \boldsymbol{e}_{d}, -\hat{p}_{1} \boldsymbol{e}_{1}, \ldots, -\hat{p}_{1} \boldsymbol{e}_{d} ]^{\!\top } \! \in \! \mathbb {R}^{(2d+1) \times d} \!
\end{equation*}
given that $\mathbb{e}_i$ is the d-dimensional standard basis vector with the i\textsuperscript{th} element being 1.


\subsubsection{FastFood\textsubscript{L}}
This (see Equation \ref{eq:fastfood}) is proposed to be learnable by \citet{yang2015carte} with two variations: FSARD (scaling matrix $\boldsymbol S$ previously sampled from chi-squared is made learnable) and FSGBARD (optimizes the marginal likelihood with respect to the diagonal matrices $\boldsymbol G$ and $\boldsymbol B$). It is further adapted by \citet{chowdhury2022learning} to make $\boldsymbol S$ or $\boldsymbol{S}, \boldsymbol{G}, \boldsymbol{B}$ learnable parameters. In our experiments, we maintain this set up.

\begin{figure}[ht]
    \centering
    \includegraphics[width=\textwidth]{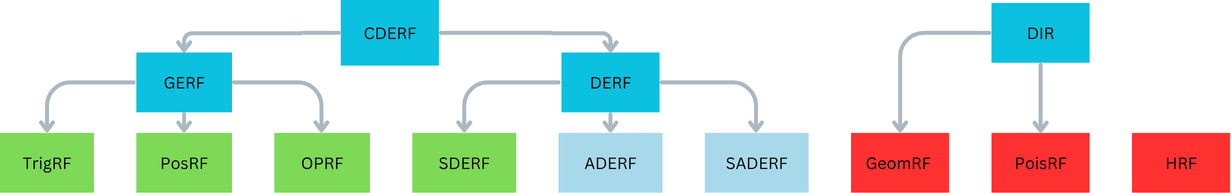}
    \caption{\textit{Spectraformer} component functions. Dark blue boxes are technique groupings. Arrow denotes which technique belongs to which grouping. Red boxes are techniques which have been explored in the literature but not in this work. Green boxes are techniques which have been explored in the literature and this work. Light blue boxes are techniques not studied empirically previously.} 
    \label{fig:component_function}
\end{figure}

\subsection{Component functions}\label{sec:method_compfunc}
\textit{Spectraformer} incorporates component functions $f$ such as PosRF, OPRF, SADERF. Component functions in \textit{Spectraformer} combine each weight matrix row $\boldsymbol{\omega}_i$ with the input $\boldsymbol{x}$. The base case of the component function is TrigRF. Component function is given in Definition \ref{definition:component_function}

\begin{definition}\label{definition:component_function}
A function of $\boldsymbol{x}$ with respect to $\boldsymbol{\omega}$ is a valid component function if and only if $\mathcal{K} \approx \phi_{\boldsymbol{\omega}}(\boldsymbol{x})^T \phi_{\boldsymbol{\omega}}(\boldsymbol{y})$ given that $\boldsymbol{\omega}_i \sim p(.)$ in Equation \ref{eq:random_features}.
\end{definition}

It follows that all component functions have provided theoretical guarantees given that $\boldsymbol{W}$ is a random matrix. In addition, PosRF, OPRF, and SADERF have also been proven theoretically to approximate $\mathcal{K}_{RBF}$ given $\boldsymbol{W}$ being the ORF.

The alternatives for component functions allow \textit{Spectraformer} to approximate the RBF with $p(.)$ corresponding to the Gaussian. \citet{likhosherstov2023dense, crt2022} show that SADERF-ORF is the SOTA combination followed by OPRF-ORF and PosRF-ORF. 

Due to page limits, we will not discuss component functions which are not included in our empirical experiments. These techniques are still included in figures and mentioned where necessary to motivate future work in the area. These include DIRF~\citep{crt2022} techniques (PoisRF, GeomRF) and HRF~\citep{choromanski2022hybrid}.


\subsubsection{CDERF}
Complex dense exponential random feature is given in Equation \ref{eq:derf}, where $\boldsymbol{\omega}, \boldsymbol{x} \in \mathbb{R}^d$, $p = \mathcal{N}(0,1)^d$, $k \in \{1, 2\}$ and $\mathbb{S}_{d_c}$ being a set of $d*d$ complex symmetric matrices. This is the general formula for most of the component functions we are interested in. The parameters and constraints of different CDERF functions are specified in Table \ref{table:derf}:
\begin{equation}\label{eq:derf}
    f_{DE}^{(k)}(\boldsymbol{\omega}, \boldsymbol{x}) = Real(D \exp(\boldsymbol{\omega}^T \boldsymbol{A} \boldsymbol{\omega} + \boldsymbol{\omega}^T \boldsymbol{B}^{(k)} \boldsymbol{x} + \boldsymbol{x}^T \boldsymbol{C}^{(k)} \boldsymbol{x}))
\end{equation}

\begin{table}[ht]
\caption{An overview of the parameters and constraints for different functions within the Complex Dense Exponential Random Feature (CDERF) family.}
\label{table:derf}

\centering
\scalebox{0.8}{
\begin{tabular}{lcccccccc}
\toprule
\textbf{}       & $\boldsymbol{A}$          & \multicolumn{1}{c}{$\boldsymbol{B}^{(1)}$}          & $\boldsymbol{B}^{(2)}$          & \multicolumn{1}{c}{$\boldsymbol{C}^{(1)}$}       & $\boldsymbol{C}^{(2)}$       & $D$          & $\boldsymbol{x}$      & $\boldsymbol{y}$           \\
\midrule
\textbf{CDERF}  & $\mathbb{S}_{d_C}$    & \multicolumn{2}{c}{$\mathbb{C}^{d*d}$}                                        & \multicolumn{2}{c}{$\mathbb{C}^{d*d}$}                                  & $\mathbb{R}$ & $\boldsymbol{x}$      & $\boldsymbol{y}$           \\
\textbf{DERF}~\citep{likhosherstov2023dense}   & $\mathbb{S}_d$        & \multicolumn{2}{c}{$\mathbb{R}^{d*d}$}                                        & \multicolumn{2}{c}{$\mathbb{R}^{d*d}$}                                  & $\mathbb{R}$ & $\boldsymbol{x}$      & $\boldsymbol{y}$           \\
\textbf{SADERF}~\citep{likhosherstov2023dense} & $A_{GE} \boldsymbol{I}_d$ & \multicolumn{2}{c}{$B_{GE} \boldsymbol{I}_d$}                                     & \multicolumn{2}{c}{$C_{GE} \boldsymbol{I}_d$}                               & $D_{GE}$     & $\Psi \boldsymbol{x}$ & $\Psi^{-1} \boldsymbol{y}$ \\
\textbf{GERF}~\citep{crt2022}   & $A_{GE} \boldsymbol{I}_d$ & \multicolumn{1}{c}{$B_{GE}^{(1)} \boldsymbol{I}_d$} & $B_{GE}^{(2)} \boldsymbol{I}_d$ & \multicolumn{2}{c}{$C_{GE} \boldsymbol{I}_d$}                               & $D_{GE}$     & $\boldsymbol{x}$      & $\boldsymbol{y}$           \\
\textbf{TrigRF}~\citep{Rahimi2007RandomFF} & $0 \boldsymbol{I}_d$      & \multicolumn{1}{c}{$i \boldsymbol{I}_d$}            & $-i \boldsymbol{I}_d$           & \multicolumn{2}{c}{$0 \boldsymbol{I}_d$}                                    & $1$          & $\boldsymbol{x}$      & $\boldsymbol{y}$           \\
\textbf{PosRF}~\citep{Choromanski2020RethinkingAW}  & $0 \boldsymbol{I}_d$      & \multicolumn{2}{c}{$1 \boldsymbol{I}_d$}                                          & \multicolumn{2}{c}{$-1 \boldsymbol{I}_d$}                                   & $1$          & $\boldsymbol{x}$      & $\boldsymbol{y}$           \\
\textbf{OPRF}~\citep{crt2022}   & $0 \boldsymbol{I}_d$      & \multicolumn{2}{c}{$1 \boldsymbol{I}_d$}                                          & \multicolumn{2}{c}{$-1 \boldsymbol{I}_d$}                                   & $1$          & $\boldsymbol{x}$      & $\boldsymbol{y}$           \\
\bottomrule
\end{tabular}
}
\end{table}

\subsubsection{TrigRF}
Trigonometric RF~\citep{Rahimi2007RandomFF} is the base RFF implementation.
\begin{equation}\label{eq:TrigRF}
\begin{array}{c}
    f_{TrigRF}^{(1)}(\boldsymbol{\omega}, \boldsymbol{x}) = \sqrt{2}\cos(\boldsymbol{\omega}^T \boldsymbol{x} + \boldsymbol{b})\\
    f_{TrigRF}^{(2)}(\boldsymbol{\omega}, \boldsymbol{y}) = \sqrt{2}\sin(\boldsymbol{\omega}^T \boldsymbol{y} + \boldsymbol{b})\\
    \boldsymbol{b} \sim Uniform(0, 2\pi)
\end{array}
\end{equation}
The use of the trigonometric sine and cosine functions leads to unstable behavior when the inputs have negative dimension-values. This can be further exacerbated when the values TrigRF try to approximate are close to 0 (since most values are of low significance). This causes the variance to approach infinity \citep{Choromanski2020RethinkingAW, crt2022}. Therefore, we do not want to use TrigRF to perform kernel approximation in attention.

\subsubsection{PosRF}
Positive RF~\citep{Choromanski2020RethinkingAW} fixes the problem of TrigRF by enforcing positive component function output in the softmax. The variance of PosRF (in contrast to the variance of TrigRF approaching infinity) approaches 0 as the approximated value of the Softmax kernel approaches 0. PosRF has two forms: Positive RF Base (PosRF-B) (see Equation \ref{eq:PosRF-B}) and Positive RF Hyperbolic (PosRF-Hyp), which is multi-component, i.e., Equation \ref{eq:rf_general_more}: $l=2$, (see Equation \ref{eq:PosRF-Hyp}).
\begin{equation}\label{eq:PosRF-B}
\begin{array}{c}
    f_{PosRF_B}(\boldsymbol{\omega}, \boldsymbol{x}) = \exp(\boldsymbol{\omega}^T \boldsymbol{x} - \frac{||\boldsymbol{x}||^2}{2})
\end{array}
\end{equation}
\begin{equation}\label{eq:PosRF-Hyp}
\begin{array}{ll}
    f_{1,PosRF_{Hyp}}(\boldsymbol{\omega}, \boldsymbol{x}) = \exp(\boldsymbol{\omega}^T \boldsymbol{x} - ||\boldsymbol{x}||^2)\\
    f_{2,PosRF_{Hyp}}(\boldsymbol{\omega}, \boldsymbol{x}) = \exp(-\boldsymbol{\omega}^T \boldsymbol{x} - ||\boldsymbol{x}||^2)
\end{array}
\end{equation}
We only use PosRF-B in our experiments.

\subsubsection{GERF}
Generalized exponential RF~\citep{crt2022} generalizes both TrigRF and PosRF with Equation \ref{eq:gerf_first_eq}.
\begin{equation}\label{eq:gerf_first_eq}
\begin{array}{c}
    f_{GERF}^{(1)}(\boldsymbol{\omega}, \boldsymbol{x}) = D \exp(A ||\boldsymbol{\omega}||^2 + B \boldsymbol{\omega}^T \boldsymbol{x} + \boldsymbol{C}||\boldsymbol{x}||^2)\\
    f_{GERF}^{(2)}(\boldsymbol{\omega}, \boldsymbol{y}) = D \exp(\boldsymbol{A} ||\boldsymbol{\omega}||^2 + sB \boldsymbol{\omega}^T \boldsymbol{y} + \boldsymbol{C}||\boldsymbol{y}||^2)\\
    \\
    Re(1 - 4A) > 0, \boldsymbol{B} = \sqrt{s(1 - 4 \boldsymbol{A})}\\
    \boldsymbol{C} = -(s+1)/2, D = (\sqrt[4]{1 - 4 \boldsymbol{A}})^d\\
    \boldsymbol{A} \in \mathbb{C}, s \in \{-1, +1\}
\end{array}
\end{equation}
where $\sqrt{.}$ and $\sqrt[n]{.}$ denoting a principal root with a complex argument

\subsubsection{OPRF}
Optimized positive RF~\citep{crt2022} is the solution to the minimization of the variance of GERF. Specifically it is defined as Equation \ref{eq:gerf_first_eq} with $s=+1$, $||\boldsymbol{x}+\boldsymbol{y}||^2 > 0$, and $\boldsymbol{A} \in \mathbb{R}$ defined in terms of $p^*$ as:
\begin{equation}
\begin{array}{cc}
    \boldsymbol{A} = (1-1/p^*)/8\\
    p^* = (\sqrt{(2||\boldsymbol{x}+\boldsymbol{y}||^2 + d)^2 + 8d||\boldsymbol{x}+\boldsymbol{y}||^2}- 2||\boldsymbol{x}+\boldsymbol{y}||^2 - d)\\/(4||\boldsymbol{x}+\boldsymbol{y}||^2)    
\end{array}
\end{equation}
Whilst PosRF is not bounded, OPRF is. OPRF can provide $e^{60}$ $\times$ variance reduction in estimating the Softmax compared to TrigRF.

\subsubsection{DERF}
Dense-exponential random features~\citep{likhosherstov2023dense} extends GERF and replace $\boldsymbol{A}, \boldsymbol{B}, \boldsymbol{C}$ with dense matrices. DERF is CDERF when $\boldsymbol{B}, C$ are in the real instead of the complex plane. Minimizing the variance of DERF leads to two approaches: ADERF and SDERF. However both these approaches rely on SVD and eigen decompositions which are not extensively supported on GPU and deep learning libraries. Therefore SADERF is proposed.

\subsubsection{SADERF}
Simplified ADERF~\citep{likhosherstov2023dense} is a special case of ADERF and extends GERF, requiring only basic unary operations in addition.
\begin{equation}\label{eq:saderf}
\begin{array}{c}
    f_{SADE}^{(1)}(\boldsymbol{\omega}, \boldsymbol{x}) = f_{GE}^{(1)}(\boldsymbol{\omega}, \Psi \boldsymbol{x}), f_{SADE}^{(2)}(\boldsymbol{\omega}, \boldsymbol{y}) = f_{GE}^{(2)}(\boldsymbol{\omega}, \Psi^{-1} \boldsymbol{y})\\
    \Psi_{l,l}^*=(\sum_j(\boldsymbol{y}_l^{(i)})^2/\sum_i(\boldsymbol{x}_l^{(i)})^2)^{1/4}\end{array}
\end{equation}

\begin{table}[h]
\caption{A summary of research gap in the literature on kernelized attention. The table shows combinations of weight matrices (columns) and component functions (rows). For each cell, the top item is the cited work for this combination for typical kernel tasks, while the bottom item is the cited work for Transformer-based tasks. $\times$ indicates that a combination has not been previously studied in either general kernel tasks or specifically for Transformers. \textit{Spectraformer} explores all listed combinations (excluding `Base' due to exhasutive pre-existing work)}
\label{table:research_gap_summary}

\centering
\newcommand{\cellstack}[2]{\begin{tabular}{@{}c@{}}#1 \\ #2\end{tabular}}

\scalebox{0.8}{
\begin{tabular}{@{}lccccccc@{}}
\toprule
\multicolumn{1}{c}{$f$} & \textbf{Base} & \textbf{ORF} & \textbf{SORF} & \textbf{QMC} & \textbf{MM} & \textbf{SGQ} & \textbf{FastFood\textsubscript{L}} \\ 
\midrule
\textbf{TrigRF} & 
\cellstack{\citep{Rahimi2007RandomFF}}{\citep{Choromanski2020RethinkingAW}} & 
\cellstack{\citep{Yu2016OrthogonalRF}}{\citep{Choromanski2020RethinkingAW}} & 
\cellstack{\citep{Yu2016OrthogonalRF}}{$\times$} & 
\cellstack{\citep{avron2016quasi}}{$\times$} & 
\cellstack{\citep{shen2017random}}{$\times$} & 
\cellstack{\citep{dao2017gaussian}}{$\times$} & 
\cellstack{\citep{yang2015carte}}{\citep{chowdhury2022learning}} \\
\addlinespace 
\textbf{PosRF} & 
\cellstack{\citep{Choromanski2020RethinkingAW}}{\citep{Choromanski2020RethinkingAW}} & 
\cellstack{\citep{Choromanski2020RethinkingAW}}{\citep{Choromanski2020RethinkingAW}} & 
$\times$ & 
$\times$ & 
$\times$ & 
$\times$ & 
\cellstack{$\times$}{\citep{chowdhury2022learning}} \\
\addlinespace
\textbf{OPRF} & 
\cellstack{\citep{crt2022}}{\citep{crt2022}} & 
\cellstack{\citep{crt2022}}{\citep{crt2022}} & 
$\times$ & 
$\times$ & 
$\times$ & 
$\times$ & 
$\times$ \\
\addlinespace
\textbf{SADERF} & 
\cellstack{\citep{likhosherstov2023dense}}{\citep{likhosherstov2023dense}} & 
\cellstack{\citep{likhosherstov2023dense}}{\citep{likhosherstov2023dense}} & 
$\times$ & 
$\times$ & 
$\times$ & 
$\times$ & 
$\times$ \\ 
\bottomrule
\end{tabular}
}
\end{table}

\subsection{Utility}
There has been an unsystematic attempt at combining component functions and weight matrices. On one hand, most component functions have had theoretical and experimental results in combining with Base and ORF. On the other hand, most weight matrices have had results in combining with TrigRF or PosRF. This means that a lot of potential combinations have not been studied previously. Table \ref{table:research_gap_summary} shows the wide research gap across 14 combinations which have never been studied in either the kernel or Transformer setting. \textit{Spectraformer} highlights this research gap. Excluding TrigRF due to its instability and Base due to its under-performance, we conduct experiments on 18 combinations.

\textit{Spectraformer} is a generic random feature framework that allows for the combination of any weight matrix with any component function provided that the weight matrix satisfies Definition \ref{definition:weight_matrix} and the component function satisfies Definition \ref{definition:component_function}. Instructions on adding new component functions or weight matrices to \textit{Spectraformer} are detailed step by step in the repository linked in the abstract. We note that \textit{Spectraformer} is an experimental rather than a theoretical framework. Aside from the challenge of generalizing the large number of weight matrices and component functions, estimating the ``actual error of the approximation, or how this error will propagate into downstream learning tasks'' is not possible \citep{error_estimation}. This is because, as \citet{error_estimation} pointed out: the RFF literature has largely use theoretical bounds which are largely impractical due to being either highly conservative or making use of ``unknown qualities'' as evident in Section \ref{sec:method}. Thus, we seek to validate its feasibility and show its ability to discover novel and performant combination via empirical results and statistical analyses.

\section{Results}\label{sec:results}
\subsection{Experimental Details}
The LRA covers tasks of different sequence lengths, difficulty, and objective, and is designed to evaluate efficient Transformers. The LRA is preferred over other benchmarks such as GLUE~\citep{glue} since efficient Transformers are designed to address tasks with long input length which reduces Transformer's training and inference speed to the default attention's quadratic time complexity. The LRA consists of ListOps~\citep{nangia-bowman-2018-listops} (testing the parsing ability of models), Text (Byte-Level Text Classification, using the IMDb review dataset)~\citep{howard-ruder-2018-universal}, Retrieval (Byte-Level Document Retrieval, using the ACL Anthology Network dataset)~\citep{retrieval}, Image (Image Classification on Sequence of Pixels in 1D projected from the 2D of images in the CIFAR-10 dataset \citep{krizhevsky2009learning}), and Pathfinder (Long-Range Spatial Dependency)~\citep{NEURIPS2018_ec895663, kimdisentangling}. We discuss the task characteristics in greater details in Section \ref{sec:dataset-char}. All our models are experimented on  NVIDIA V100 and 12 CPU with 20GB of memory. Our codebase is based on and includes the adapted or original implementation from \citep{chen2021skyformer, chowdhury2022learning, liu2021random, Choromanski2020RethinkingAW, crt2022, likhosherstov2023dense, wang2020linformer, nystromformer}, \citep{kitaev2020reformer, hadamard, informer} (Apache License, Version 2.0).


The parameters are identical to \citep{chen2021skyformer} and chosen to limit the parameter count to account for the marked training time of multiple models. All the parameters for \textit{Spectraformer} variants are kept identical for fair comparison, with the only difference being the number of random feature. We vary the number of random feature between 64, 128, 256, and 512 for comprehensive analysis. Dimension as 128 was used in \citep{chen2021skyformer} and 256 was used in \citep{Choromanski2020RethinkingAW, crt2022, likhosherstov2023dense}. Our hyperparameters are available in Table \ref{table:hyperparams}. Our code is implemented in Python 3.12 and Pytorch, we use the Transformer base from \citep{chen2021skyformer}. The model name convention is [component function]-[weight matrix] (e.g., \textit{OPRF-FastFood\textsubscript{L}}).

\subsection{Overall Results}
Our comprehensive experiments on the Long Range Arena (LRA) benchmark reveal that the \textit{Spectraformer} framework not only produces highly competitive models but also establishes a new state-of-the-art for random feature-based efficient Transformers. Before diving into a detailed breakdown, we first present a high-level comparison of our top-performing variants against a wide range of established efficient Transformers in Table \ref{table:first_table}. This includes methods based on downsampling (Nystromformer~\citep{nystromformer}, Skyformer~\citep{chen2021skyformer}, BigBird~\citep{zaheer2020big}), learnable patterns (Reformer~\citep{kitaev2020reformer}), and other linearization approaches (Linformer~\citep{wang2020linformer}).

As shown in Table 1, our best model, \textit{OPRF-FastFood\textsubscript{L}}, achieves a mean accuracy of 57.17\%, which is highly competitive with leading methods like Nystromformer~\citep{nystromformer} (58.06\%) and BigBird~\citep{zaheer2020big} (57.25\%). This is a significant finding, as it demonstrates for the first time that a well-configured kernelized attention model can perform at the same level as other classes of efficient Transformers on this challenging benchmark. Notably, \textit{OPRF-FastFood\textsubscript{L}} is on par with BigBird~\citep{zaheer2020big} on mean accuracy while offering a substantial reduction in peak memory consumption (4.65 to 1.82 GB) and training time (3.32 hours to 2.26 hours)

To provide full transparency and detail on how our top variants were identified, Table \ref{table:lra_1} presents the complete results for all 18 combinations of component functions and weight matrices explored within \textit{Spectraformer}, tested with random feature dimensions of 64, 128, 256, and 512. The previous state-of-the-art random feature models, OPRF-ORF and SADERF-ORF, are highlighted in bold for direct comparison.

The results in Table \ref{table:summary_stats} confirm that several novel combinations discovered through our framework, such as OPRF-FastFood\textsubscript{L} and PosRF-MM, consistently outperform the previous random feature SOTAs. To help navigate these extensive results, Table \ref{table:best-model-based-on-category} summarises the best-performing model for each task according to accuracy, training time, and memory usage. This highlights the framework's ability to generate variants that offer different trade-offs, catering to specific computational or performance requirements. We now proceed to a deeper analysis of these results.

\begin{table}[h]
\caption{A summary of the best-performing \textit{Spectraformer} models, with tasks as columns and optimization categories as rows. Results are shown for 64, 128, 256, and 512 feature settings.}
\label{table:best-model-based-on-category}
\centering
\resizebox{\linewidth}{!}{
\begin{tabular}{@{}ll cccccc@{}}
\toprule
\textbf{Metric} & \textbf{Feat} & \multicolumn{6}{c}{\textbf{Task}} \\
\cmidrule(lr){3-8}
 &  & \textbf{L} & \textbf{T} & \textbf{R} & \textbf{I} & \textbf{P} & \textbf{$\mu$} \\
\midrule
\multirow{4}{*}{\textbf{Accuracy}}
& 64 & OPRF-QMC & OPRF-SORF & OPRF-MM & SADERF-FastFoodL & PosRF-QMC & OPRF-FastFoodL \\
& 128 & OPRF-MM & SADERF-SORF & OPRF-ORF & OPRF-FastFoodL & OPRF-SGQ & OPRF-FastFoodL \\
& 256 & OPRF-FastFoodL & SADERF-SORF & SADERF-ORF & OPRF-FastFoodL & OPRF-FastFoodL & OPRF-FastFoodL \\
& 512 & OPRF-FastFoodL & SADERF-SORF & OPRF-MM & OPRF-FastFoodL & PosRF-ORF & OPRF-FastFoodL \\
\midrule
\multirow{4}{*}{\textbf{Time}}
& 64 & PosRF-MM & PosRF-QMC & PosRF-SGQ & PosRF-MM & PosRF-MM & PosRF-MM \\
& 128 & PosRF-SORF & PosRF-QMC & PosRF-QMC & PosRF-QMC & PosRF-SORF & PosRF-QMC \\
& 256 & PosRF-MM & PosRF-SORF & PosRF-MM & PosRF-SGQ & PosRF-QMC & PosRF-MM \\
& 512 & \multicolumn{6}{c}{PosRF-FastFoodL} \\
\midrule
\multirow{4}{*}{\textbf{Memory}}
& 64 & \multicolumn{6}{c}{PosRF-MM} \\
& 128 & \multicolumn{6}{c}{PosRF-FastFoodL} \\
& 256 & \multicolumn{6}{c}{PosRF-FastFoodL} \\
& 512 & \multicolumn{6}{c}{PosRF-FastFoodL} \\
\bottomrule
\end{tabular}
}
\end{table}

\subsection{Task Characteristics}\label{sec:dataset-char}

The tasks in the LRA include: ListOps~\citep{nangia-bowman-2018-listops}, Text (Byte-Level Text Classification, using the IMDb review dataset)~\citep{howard-ruder-2018-universal}, Retrieval (Byte-Level Document Retrieval, using the ACL Anthology Network dataset)~\citep{retrieval}, Image (Image Classification), and Pathfinder (Long-Range Spatial Dependency)~\citep{NEURIPS2018_ec895663, kimdisentangling}. We take into consideration three main characteristics:
\begin{itemize}
    \item \textbf{Number of output classes}. ListOps is a task which inputs a sequence containing brackets and simple mathematical operators, with the output being a integer between 1 and 10, hence it has 10 classes. Text is a sentiment classification task on an IMDb review dataset with the class being either negative or positive, hence it has 2 classes. Retrieval inputs two sequences and the output is either False (meaning the sequences are not of the same document) or True (meaning the sequences are of the same document). Image is based on the CIFAR-10~ \citep{krizhevsky2009learning} and has 10 classes. Pathfinder~\citep{NEURIPS2018_ec895663, kimdisentangling} has 2 classes and requires the model to identify the target contour between one end and the other.
    \item \textbf{Maximum sequence size}. ListOps is 2K; Text is 4K; whilst Retrieval requires the processing of two sequences each of 4K length making the total maximum sequence size being 8K; Image flattens the 2D of CIFAR-10 images of 32-by-32 into a list with the size of 1024; Similary, Pathfinder also takes in a 1024-dimensional list from 32-by-32 images.
    \item \textbf{Required attention span}. This is the `mean distance between the query token and the attended tokens, scaled by attention weights'. The average values for required attention span calculated by \citep{tay2021long} is: ListOps being 0.7589, Text being 0.3194, Retrieval being 1.329, Image being 0.359 Pathfinder being 0.5371.
\end{itemize}

\subsection{Analysis of Methods}\label{sec:analysis_of_methods}

\begin{figure}[h]
    \centering
    \includegraphics[width=\textwidth]{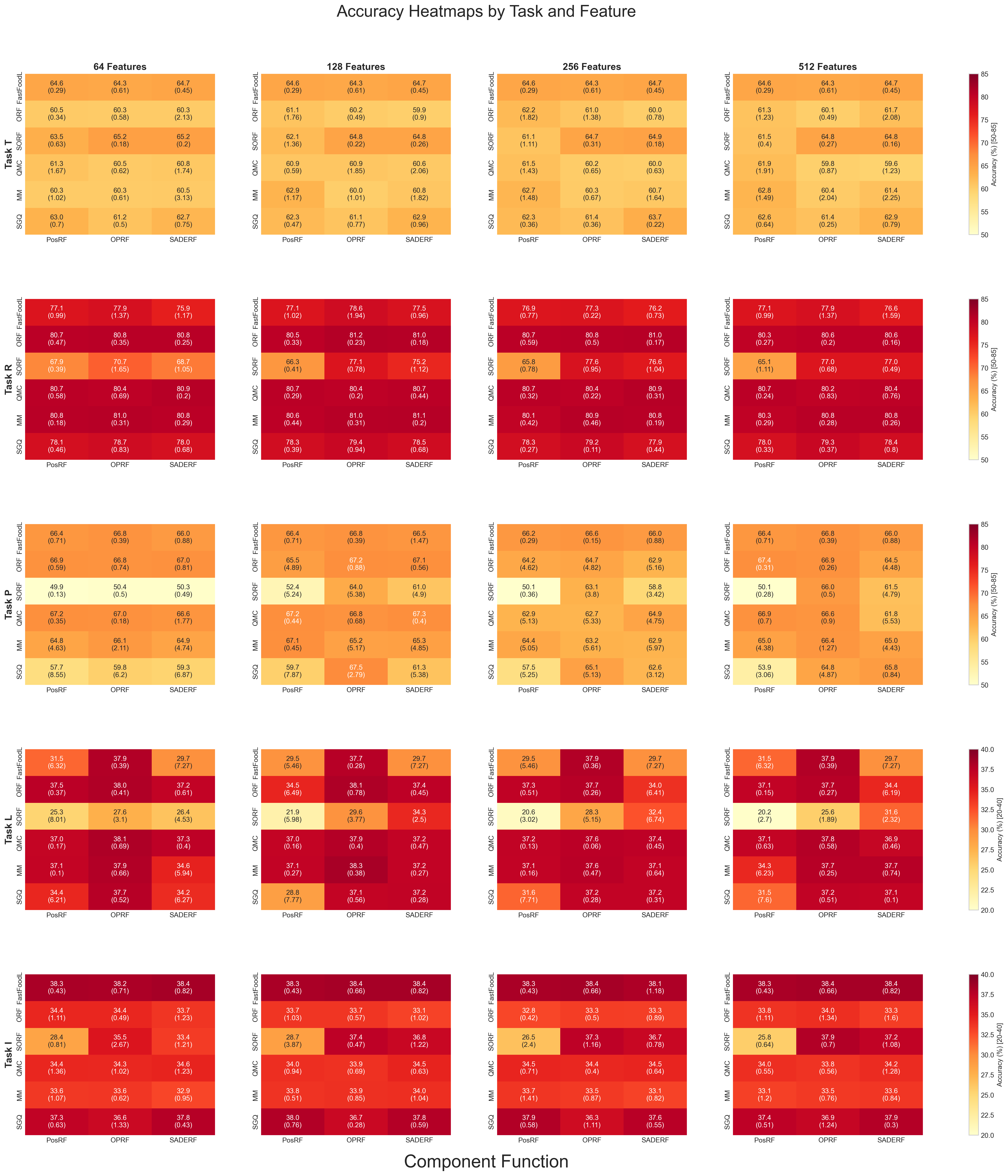} %
\caption{Heatmaps of \textit{Spectraformer} combinations for the 64, 128, 256, 512 features. $L$ and $I$ have 10 classes, opposed to $T$, $R$, $P$ having 2 classes. Hence we adjust the vmin-vmax for the first group as 20-40 with the second group as 50-85.}
\label{fig:heatmap}
\end{figure}

\begin{table}[t]
    \centering
    \caption{Pearson correlation coefficients between model performance metrics (Mtr.) (Accuracy (Acc), Time (Time), Memory (Mem)) and task characteristics. Task characteristics include the number of classes (Cl.), input sequence length (Len.), and required attention span (Att.). The analysis is presented for component functions (left) and weight matrices (right).}
    \label{table:pearson}
    \resizebox{0.8\textwidth}{!}{%
        \begin{tabular}{llccccccccc}
            \toprule
            \multicolumn{2}{c}{} & \multicolumn{3}{c}{$f$} & \multicolumn{6}{c}{$W$} \\
            \cmidrule(lr){3-5} \cmidrule(lr){6-11}
            \textbf{Mtr.} & \textbf{T} & \textbf{PosRF} & \textbf{OPRF} & \textbf{SADERF} & \textbf{FastFood\textsubscript{L}} & \textbf{QMC} & \textbf{MM} & \textbf{ORF} & \textbf{SGQ} & \textbf{SORF} \\
            \midrule
            \multirow{3}{*}{Acc} & Cl. & -0.908 & -0.924 & -0.921 & -0.950 & -0.921 & -0.919 & -0.921 & -0.905 & -0.913 \\
            & Len. & 0.706 & 0.720 & 0.722 & 0.664 & 0.735 & 0.741 & 0.728 & 0.754 & 0.696 \\
            & Att. & 0.459 & 0.516 & 0.494 & 0.404 & 0.566 & 0.562 & 0.557 & 0.506 & 0.356 \\
            \cmidrule(l){2-11}
            \multirow{3}{*}{Time} & Cl. & 0.169 & 0.191 & 0.185 & 0.234 & 0.176 & 0.177 & 0.165 & 0.177 & 0.176 \\
            & Len. & -0.169 & -0.198 & -0.186 & -0.215 & -0.180 & -0.180 & -0.175 & -0.184 & -0.180 \\
            & Att. & -0.271 & -0.275 & -0.265 & -0.364 & -0.255 & -0.255 & -0.252 & -0.259 & -0.255 \\
            \cmidrule(l){2-11}
            \multirow{3}{*}{Mem} & Cl. & 0.160 & 0.144 & 0.143 & 0.241 & 0.160 & 0.160 & 0.149 & 0.161 & 0.160 \\
            & Len. & -0.185 & -0.157 & -0.159 & -0.253 & -0.177 & -0.177 & -0.171 & -0.181 & -0.177 \\
            & Att. & -0.265 & -0.233 & -0.238 & -0.399 & -0.259 & -0.259 & -0.255 & -0.263 & -0.259 \\
            \bottomrule
        \end{tabular}
    }
\end{table}

We now explore the suitability of weight matrices and component functions depending on task characteristics. Table \ref{table:summary_stats} shows the mean statistics across component functions and weight matrices for each task as well as average performance across tasks. To ensure the performance of methods consistently across different feature settings, we also refer to Figure \ref{fig:heatmap}.

First, we will analyze the component functions and weight matrices in terms of accuracy. For component functions, we can see that OPRF and SADERF outperform PosRF for all tasks except Text. For weight matrices, we observe the following:

\begin{itemize}
    \item $T$: FastFood\textsubscript{L} and SORF performs well in $T$ but other models also perform equally well in task $T$. This task is not too hard.
    \item $R$: ORF, QMC, and MM perform the best. FastFood\textsubscript{L} and SGQ perform well moderately, and SORF performs the worst.
    \item $P$: Most models perform well (ORF, FastFood\textsubscript{L}, QMC, MM). SGQ slightly less well than this group. SORF performs quite poorly. 
    \item $L$: QMC and MM perform the best. ORF performs moderately well. SORF performs the worst. Meanwhile, SGQ's performance varies with respect to the number of random features.
    \item $I$: FastFood\textsubscript{L} performs the best, followed by SGQ. ORF, QMC and MM perform moderately. With SORF performing the worst.
\end{itemize}


Based on the above observations, we note that ORF, QMC, and MM tend to have the same trend. These weight matrices along FastFood\textsubscript{L} (albeit having a different trend) have high mean performance (see Table \ref{table:summary_stats}) and high performance per task. This is followed by SGQ. SORF has the worst performance. QMC and MM having the same trend is expected since MM is an improvement on QMC using the same base formula and assumption. Empirically, QMC also outperforms ORF marginally as expected, based on existing work (see \citep{liu2021random}, Table 9 (Appendix) w.r.t. Gaussian training error). QMC and MM are based on Equation \ref{eq:unit_cube} which is a reformulation of Equation \ref{eq:bochner} whilst reducing variance. Thus, theoretically QMC and MM should behave similarly yet better than MC-based methods like ORF. SORF is the faster version of ORF \citep{Yu2016OrthogonalRF}, and empirically it becomes nearly unbiased with the number of features above 32, essentially collapsing to the bias of ORF. Yet in practice, \citet{liu2021random} show that SORF does not perform as well as ORF in classification tasks, and this is reflected in our experiments. In addition, \citet{liu2021random} also show that FastFoodF (which is the FastFood approximator (FastFood fixed) as opposed to FastFood\textsubscript{L}, the FastFood learner) performs as well as QMC. Based on \citep{chowdhury2022learning} however, it is expected that FastFood\textsubscript{L} exceeds this. And being the weight learner class, its behaviors deviate from the general trend of weight matrices. \citet{liu2021random} show empirically that GQ performs as well as QMC. However, GQ cannot be used in practice, hence the use of SGQ. It was also not compared empirically in \citep{liu2021random}. Given that SGQ approximates a kernel $k$ that is subgaussian with parameter $b$, $A \geq 24eb^2M^2$, $d$ being the dimension of the data, $M \subset \mathbf{R}^d$ be the diameter between inputs $x$ and $y$, then SGQ's sampling complexity is bounded by $2^d \left( \frac{12eb^2M^2}{A} \right)^A$.

This means that the dimension of the token as well as the distance between two tokens being calculated will increase the complexity. Since our dimension of the token is quite high compared for most kernel tasks (64), and the distance between two tokens is quite high as well due to the tasks being from the LRA, this results in SGQ's rather average performance. The relationship between the distance of the tokens and the approximation quality of SGQ is apparent in the moderate Pearson correlation coefficient of 0.506 between SGQ and attention span in Table \ref{table:pearson}. Unfortunately, we have less than an obvious idea for the relationship between the other weight matrices and token characteristics like SGQ since most of their sampling complexity is only dependent on the number of features \citep{liu2021random}.


Table \ref{table:pearson} shows that, as expected, the number of classes is the primary determinant of models' performance. However, the input length also plays a role in determining the outcome of the performance of a weight matrix in a task as well. We observe a similar trend across ORF, QMC, and MM as stated above across correlation coefficients between accuracy w.r.t. number of classes, input length, and attention span. SGQ as suggested above performs similarly for the first two metrics but with a slightly lower attention span coefficient. FastFood has the lowest correlation coefficients for both input length and attention span, suggesting that its learnable weights behave contrary to task characteristics and learning might be attributed to this.

\begin{figure}[ht]
    \centering
    \includegraphics[width=\linewidth]{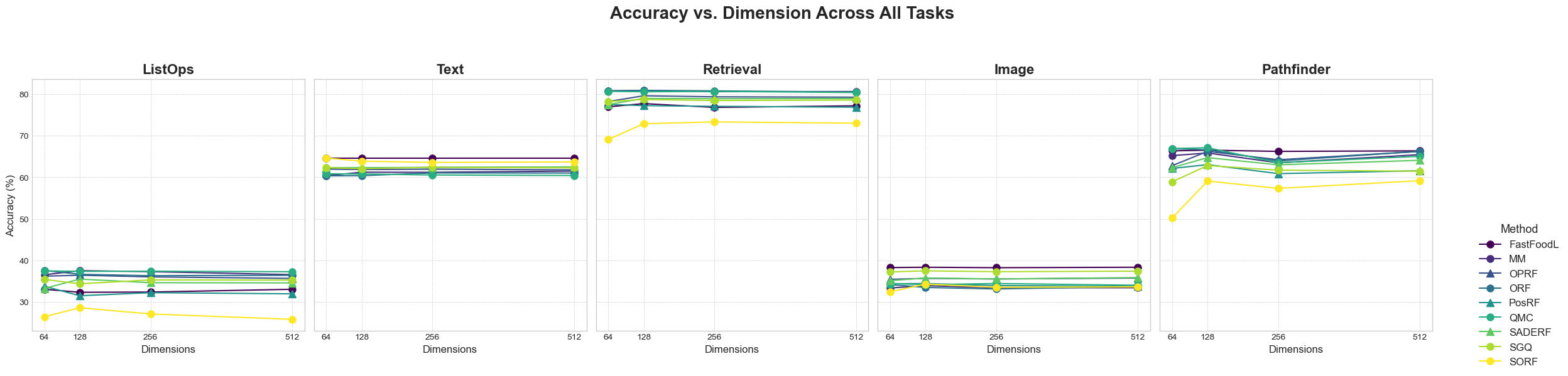}
    \caption{Accuracy vs Number of Features Across All Tasks}
    \label{fig:feat_comparison}
\end{figure}

Figure \ref{fig:feat_comparison} shows the accuracy trend for different number of features for each task. We observe that accuracy peaks mostly at 128 and sometimes 256 number of features, thereafter maintaining or decreasingly stably. We also find that there is no relationship between input length and the number of features (also see Figure \ref{fig:dimension_comparison_comprehensive}).

This suggests that increasing the number of random features past 128 induces overfitting and 128 is the ideal number of random features. The best way to confirm whether overfitting is occurring is to look at the gap between the generalization accuracy (the accuracy on the development set) and the training accuracy. We term this accuracy as the generalization gap. We calculate the average generalization gap and averaging them for all seeds of the same task and model between 512 - 256, 256 - 128, and 128 - 64 features. The result is respectively 0.143\%, 0.041\%, -0.404\%. This confirms our initial intuition that 64 features are too few (underfitting) but 256 features are too many (overfitting). The accuracy obtained is obtained with early stopping, accuracy obtained without early stopping can induce a much larger generalization gap difference. This confirms that overfitting occurs past 128 features. In addition, during experimentation, we also found that the 512 features is highly unstable due to gradient explosion from large matrix calculation leading to NaN values. This occurred for three seed (twice for PosRF-SGQ and once for SAERF-ORF).

SORF in particular performs poorly at 64 number of features with accuracy improving drastically at 128 number of features. This validates previous findings that in transformer applications, 128 features is the optimal option.

\subsection{Efficiency Analysis}

\begin{figure}[ht]
    \centering
    \includegraphics[width=\linewidth]{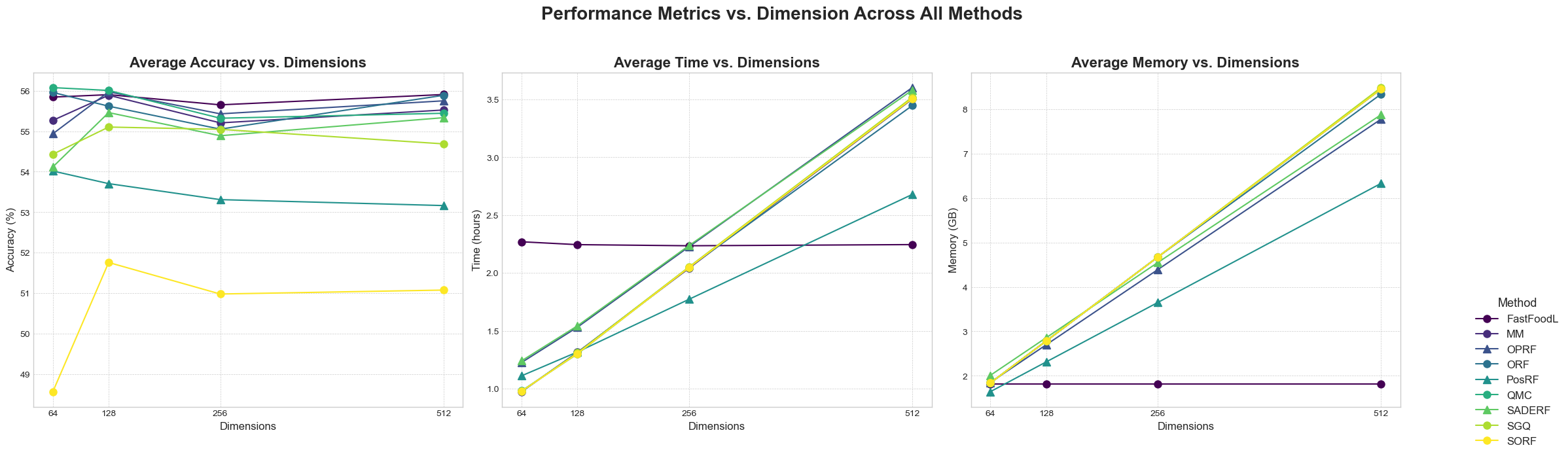}
    \caption{Performance Metrics vs. Number of Features}
    \label{fig:dimension_comparison_comprehensive}
\end{figure}

In analyzing efficiency, we follow the definition of efficient Transformers in \citep{tay_survey} in terms of memory and computation, the latter interpreted as computational cost. The overall trend of accuracy, time, and memory of component functions is that they increase linearly from PosRF, OPRF, to SADERF. Mean accuracy is respectively 53.71, 55.97 and 55.46. Mean time is respectively 1.32, 1.53 and 1.54. Mean memory is 2.31, 2.70, 2.86. Please refer to Table \ref{table:summary_stats}. These figures are for 128 random features but with other number of features, the results follow the same trend. SADERF and OPRF having a higher computational requirement than PosRF is due to them having additional computation of the L2 norm of the weight matrix which PosRF does not calculate. Hence, when limited computational resource is a concern, PosRF is a better option. Figure \ref{fig:dimension_comparison_comprehensive} shows that PosRF's training tmie and memory consumption grows at a significantly lower rate than other methods. Otherwise, OPRF provides the best accuracy. We observe the increase in training time and peak memory consumption as expected due to increasing the number of random features sampled for weight matrices (see Figure \ref{fig:dimension_comparison_comprehensive}). Training time and peak memory consumption relies more heavily on component functions than weight matrices, with the exception of FastFood\textsubscript{L}. FastFood\textsubscript{L} takes significantly less memory consumption at the expense of training time. However, whilst most methods increase training time and memory consumption significantly as the number of features increases, FastFood\textsubscript{L} remains constant (see Figure \ref{fig:dimension_comparison_comprehensive}).

In addition, Table \ref{table:pearson} shows that, in terms of weight matrices, training time and peak memory consumption is relatively stable and does not vary with respect to number of classes, input length, and the required attention span, with the exception for FastFood\textsubscript{L}. This is largely due to FastFood\textsubscript{L} modifying the attention matrix instead of approximating it. However, we do notice that there is a higher negative correlation between the required attention span and peak memory consumption.

We also analyze efficiency in terms of the number of features. Figure \ref{fig:efficiency} shows that memory usage decrease gradually as the sequence length increases and the training time remains relatively stable.



\subsection{Inductive Bias of Efficient Attention}
In addition, we also notice that, the top performing models (in Table \ref{table:first_table}) such as \textit{Spectraformer} variants, Nystromformer~\cite{nystromformer}, Big Bird~\cite{zaheer2020big}, Reformer~\cite{kitaev2020reformer}, Skyformer~\cite{chen2021skyformer} are efficient Transformers that  outperform the OT. They can be grouped into the following two induction biases:
\begin{itemize}
    \item \textbf{Low-rank bias} (\textit{Spectraformer}, Nystromformer~\cite{nystromformer}, Skyformer~\cite{chen2021skyformer}) assumes that the true attention matrix is simple, or low rank, in nature.
    \item \textbf{Sparsity bias} (Big Bird~\cite{zaheer2020big}, Reformer~\cite{kitaev2020reformer}) assumes that only a small number of attention values in the attention matrix contains meaningful information, ignoring the computation of a significant number of token-to-token attention values, since they only contribute to noise.
\end{itemize}

Efficient transformers do not just reduce the computational cost, but they also introduce strong induction bias about the structure of the attention matrix. Efficient transformers' approximation of the OT's true attention matrix should imply that their performance be upperbounded by the OT's performance. However, as we have seen in Table \ref{table:first_table}, they consistently outperform the OT, implying that at least for the characteristics of the LRA, a low-rank bias or a sparsity bias, acting as regularizers on the attention matrix, are helpful induction biases. This aligns with the analysis done in Section \ref{sec:analysis_of_methods}. 128 features will definitely produce a lower rank matrix than 256, 512 features hence models with an even lower-rank bias. This however questions why the OT outperforms 512 and 256-feature models. 128, 256 and 512 feature models have lower rank than the OT and they necessarily produce noise due to the attention approximation process. Having more features likely enlarge the noise in the final approximated attention matrix, making these models fit to spurious noise. Having too few features (64), however, underfits and lead to inexact kernel approximation.

\subsection{Method Recommendation}

Based on the above experiments, our recommendation is to avoid SORF. SORF performs well on the Text task. However, this is not captured by our metrics (number of output classes, maximum input length, and attention span). There are potential task characteristics that we cannot account for with this exception. Overall, we suggest FastFood\textsubscript{L} and QMC among the main weight matrix group of QMC, MM and ORF. They both are highly performant. FastFood\textsubscript{L} offers the advantage of low memory consumption at the cost of longer training time, and QMC maintains the average training time and memory consumption.

Our results establish \textit{OPRF-FastFood\textsubscript{L}} as the new state-of-the-art for random feature-based efficient Transformers. More importantly, our findings demonstrate that this class of models is now highly competitive with top-performing methods from other categories like Nystromformer~\citep{nystromformer} and BigBird~\citep{zaheer2020big}. Specifically, \textit{OPRF-FastFood\textsubscript{L}} achieves accuracy on par with BigBird~\citep{zaheer2020big} (57.17\% vs. 57.25\%) while offering substantial reductions in both peak memory consumption (1.82 GB vs. 4.65 GB) and training time. While Nystromformer attains a higher average accuracy (58.06\%), our model provides a valuable alternative with greater memory efficiency (1.82 GB vs. 2.63 GB). These results introduce new, valuable points to the accuracy-memory-time Pareto frontier, offering practitioners compelling options based on their specific performance and resource constraints. In addition to our top-performing model, the \textit{Spectraformer} framework also produced other novel combinations like OPRF-SGQ and PosRF-MM, which outperform previous RF-based SOTAs (OPRF-ORF and SADERF-ORF) and provide a range of options with different computational profiles. The result mentioned here is available in Table \ref{table:first_table}.

\section{Related Work}\label{appendix:related_work}
Spectraformer is based on several theoretical assumptions various mathematical frameworks. This section discusses alternatives to these assumptions. Specifically, Section \ref{appendix:kernelized_attention} discusses alternative kernelized formulations to Section \ref{sec:attention}; Section \ref{appendix:kernel_approx} discusses alternative kernel approximation techniques to random features in Section \ref{appendix:kernel_approx}. 

\subsection{Alternative Kernelized Attentions}\label{appendix:kernelized_attention}
In addition to the kernelized formulation of attention as shown in Section \ref{sec:attention}, based on our literature survey, we have identified the following three alternative kernel-based formulations of attention.

\textbf{Nadaraya-Watson kernel estimator in integration form~\citep{nguyen2022fourier}:} Given the Gaussian kernel, where the probability of a single variable and the joint probability of two variables are estimated using kernel density estimation with the Gaussian kernel, then we have Equation \ref{eq:nadaraya_int_kernel}, leading to the formulation of FourierFormer.
\begin{equation}\label{eq:nadaraya_int_kernel}
\begin{array}{ll}
\phi(\boldsymbol{\delta}) & =exp(-||\boldsymbol{\delta}||^2/2\sigma^2)\\ \hat{r}_{n}(\boldsymbol{k}) & =\int\frac{\boldsymbol{v} p(\boldsymbol{v},\boldsymbol{k})}{p(\boldsymbol{k})}dv  =\frac{\sum_{j=1}^N \hat{\boldsymbol{v}}_j \phi(\boldsymbol{k}-\boldsymbol{k}_j)}{\sum_{j=1}^N \phi(\boldsymbol{k}-\boldsymbol{k}_j)}\\ \hat{r}_{n}(\boldsymbol{q}_i) & =\sum_{j=1}^{N}softmax(\boldsymbol{q}_i^{T}\boldsymbol{k}_j/\sigma^2)\hat{\boldsymbol{v}}_{j}
\end{array}
\end{equation}

\textbf{Gaussian decomposition~\citep{song2021implicit}:} the attention mechanism can also be decomposed into the Gaussian kernel directly from Equation \ref{eq:attention-row-kernel} without appealing to the use of kernel estimator \citep{song2021implicit}. Equation \ref{eq:ika} leads to Implicit Kernel Attention (IKA).
\begin{equation}\label{eq:ika}
A_{i} = \sum_{j=1}^{N}\frac{1}{Z_l(\boldsymbol{q}_i, \boldsymbol{K})}exp(\frac{-||\boldsymbol{q}_i - \boldsymbol{k}_j||^2_2}{2 \sqrt{d_k}}) exp(\frac{||\boldsymbol{q}_i||^2_{p=2} + ||\boldsymbol{k}_j||^2_{p=2}}{2 \sqrt{d_k}})\boldsymbol{v}_j
\end{equation}

\textbf{Non-local operation~\citep{Wang2017NonlocalNN}} is popular decomposition of attention in computer vision. This is the basis for Vision Transformer and related work. Non-local operation itself a generalisation of the non-local means used in image denoising \citep{Buades2005ANA}. Non-local operation is essentially an operation that operates on the entire input range, instead of a specific window which is characteristic of convolutional operations (see Equation \ref{eq:non-local}). Indeed, the attention vector output $\boldsymbol{y}$ can be defined using $f$ which is the softmax (see Equation \ref{eq:non-local-softmax}) in the non-local operation.
\begin{equation}\label{eq:non-local}
\mathbf{\boldsymbol{y}}_{i}=\frac{1}{C(\boldsymbol{x})}\sum_{\forall j}f(\boldsymbol{x}_i,\boldsymbol{x}_j)g(\boldsymbol{x}_j)
\end{equation}
\begin{equation}\label{eq:non-local-softmax}
\mathbf{\boldsymbol{y}}=softmax(\boldsymbol{x}^{T}\boldsymbol{W}_{\theta}^{T}\boldsymbol{W}_{\theta}\boldsymbol{x})g(\boldsymbol{x})
\end{equation}

We will show that the non-local operation in Equation \ref{eq:non-local} is a generalisation of the Nadaraya-Watson kernel estimator. Specifically, when defining $C(\boldsymbol{x}) = \sum_l \mathcal{K}(\boldsymbol{q}_i, \boldsymbol{k}_l), f = \mathcal{K}(\boldsymbol{q}_i, \boldsymbol{k}_l), g = \boldsymbol{v}_j$, then we obtain the original Nadaraya-Watson kernel estimator in Equation \ref{eq:attention-row-kernel}. The variety of kernel decompositions of attention demonstrate the robustness of the kernel interpretation of attention and hence opening up an alley in attention improvement through the lens of kernel.

\subsection{Kernel Approximation Beyond Random Features}\label{appendix:kernel_approx}
Kernel approximation is presented in Section \ref{sec:random-features} through the use of random features based on Bochner's theorem limiting to only symmetric, positive-definite (PD) kernels. While Spectraformer focuses on approximating the symmetric softmax kernel, other research has extended the RFF paradigm to handle other classes of functions.

Notably, the AsK-RFFs (Asymmetric Kernel RFFs)~\citep{10.1007/s10994-024-06626-8} method generalizes RFFs to handle asymmetric kernels. The key theoretical innovation is the use of a complex measure expanding on the Bochner theorem, which can be decomposed into four finite positive measures. This allows AsK-RFFs to create two distinct feature mappings that can approximate an asymmetric kernel. A different class of kernel functions are the Generalized Zonal Kernels (GZKs) which are dot-product kernels on the unit sphere, which is different from shift-invariant kernels which are the focal point of this paper. Nevertheless, the Gaussian is a part of this class. Random Gegenbauer Features~\citep{pmlr-v162-han22g} are designed to approximate the GZKs. This is done by performing a low-rank approximation via sampling the GZK's feature map, defined by an infinite series of Gegenbauer polynomials instead of using the entire feature map.

In addition, there has been other kernel approximation techniques outside of random features. These include greedy basis selection \citep{Smola2000SparseGM}, divide and conquer \citep{Hsieh2013ADS,Zhang2013DivideAC}, nyström methods \citep{Williams2000UsingTN} (which led to the development of Skyformer~\citep{chen2021skyformer} and Nyströmformer~\citep{nystromformer}). Nystrom methods are a type of low rank attention of which Low-Rank Transformer~\citep{9053878} is a part of. There is also sparse attention (Big Bird~\citep{zaheer2020big} which combines windowed, global and random attention, Reformer~\citep{kitaev2020reformer} which uses locality-sensitive hashing, Longformer~\citep{beltagy2020longformerlongdocumenttransformer} which combines band attention and global attention on special tokens like [CLS], ETC (Encoding Long and Structured Inputs)~\citep{ainslie-etal-2020-etc}, Routing Transformer~\citep{roy-etal-2021-efficient}, Star Transformer~\citep{guo-etal-2019-star}), time-series inspired methods (Informer~\citep{informer}), or Linformer~\citep{wang2020linformer}. The inclusion of many of these models in our LRA benchmark comparison (Table \ref{table:first_table}) provides a broader comparison for evaluating the performance of \textit{Spectraformer} against other efficient Transformers beyond random features. For a more comprehensive overview, please see the survey in \citep{10322921}.

An alternative to using random features is directly specifying $\phi(x)$ without approximating the softmax via random features. cosFormer~\citep{qin2022cosformerrethinkingsoftmaxattention} specifies $\phi(x)$ as \texttt{LeakyReLU} and \texttt{ReLU}.

\section{Conclusion}\label{sec:conclusion}
We present Spectraformer, an open-source framework for approximating and learning the attention kernel in the Transformer. Our paper generalizes past works, and presents empirical findings on different component function and weight matrix combinations. We experiment with 18 combinations, of which 14 are novel. Spectraformer has produced variants that are competitive against a wide range of efficient Transformers, including those based on sparse attention and other approximation methods as shown in our expanded LRA benchmark. We have demonstrated that a novel combination discovered through our framework, OPRF-FastFoodL, performs on par with leading models like Nystromformer and BigBird, establishing a new SOTA among random feature-based methods and a strong contender in the overall efficient Transformer landscape. We suggest the use of weight matrices FastFoodL and QMC combined with component functions based on specific computational requirements (PosRF for less and OPRF for more computational requirement). Our work shows the validity of Spectraformer for studying and producing novel variants based on existing and future research in random features and distribution sampling.

\section{Future work, limitations, and broader impact}\label{sec:limitations}
Spectraformer provides an experimental framework for random features in efficient Transformer with kernelized attention. We are aware that many works cited do not provide comparable error approximation bounds. In addition, as \citep{error_estimation} point out, most theoretical bounds are either highly conservative or involving unknown qualities, therefore making comparisons impossible. Therefore, we have validated our framework on extensive tasks, and provided comprehensive analyses and intuitive explanation. We welcome future work to generalize and build on Spectraformer. We have experimented with a significant number of combinations but not the exhaustive list in Sections \ref{sec:method_weight} and \ref{sec:method_compfunc}. Future experiments can work with these additional techniques. In addition, providing a definitive theoretical reason for the success of specific method-task combinations, such as PosRF-QMC on ListOps, remains a significant open question that aligns with broader challenges in the kernel and random features literature. A rigorous theoretical investigation into these interactions is a valuable direction for future work.


Transformer-based architecture can also benefit from Spectraformer. Past random feature Transformer works \citep{crt2022,likhosherstov2023dense} have experimented with conformer and vision Transformer with ViT. \citet{Jeevan_2022_WACV} provide a good foundation to extend our work. Spectraformer, although currently experimented in the Transformer setting, can also extend to non-parametric kernel classification. It could also benefit from ablation studies like the comparison variance in \citep{chowdhury2022learning} or in \citep{error_estimation}. Given the complex nature of language models in the ethical, energy, and societal domains, our work should be used responsibly \citep{weidinger2021ethical}.

\section*{Acknowledgments}
The production of the code used in this paper was partially reliant on generative AI code assistant. All AI-generation has been validated by the authors. This research was undertaken with the assistance of resources and services from the National Computational Infrastructure (NCI), which is supported by the Australian Government. We would also like to acknowledge the ARC Centre of Excellence for Automated Decision Making and Society (CE200100005).


\bibliographystyle{ACM-Reference-Format}
\bibliography{refs}

\appendix
\newpage
\section*{Appendix}

\begin{table}[h]
\caption{Hyper-parameters for all \textit{Spectraformer} experiments on the Long Range Arena (LRA) benchmark tasks. The settings follow \citet{chen2021skyformer} to ensure a fair comparison under limited computational power.}

\label{table:hyperparams}

\centering
\scalebox{0.7}{
\begin{tabular}{lccccc}
\toprule
                      & \multicolumn{1}{c}{\textbf{ListOps (L)}} & \multicolumn{1}{c}{\textbf{Text (T)}} & \textbf{Retrieval (R)} & \textbf{Image (I)} & \textbf{Pathfinder (P)}\\
\midrule
Embedding dim.        & \multicolumn{5}{c}{64}  \\
Transformer   dim     & \multicolumn{5}{c}{64}  \\
Hidden dim            & \multicolumn{5}{c}{128} \\
Head dim              & \multicolumn{5}{c}{32}  \\
Num. heads            & \multicolumn{5}{c}{2}   \\
Num. layers           & \multicolumn{5}{c}{2}   \\
Vocabulary   size     & \multicolumn{1}{c}{32}         & \multicolumn{1}{c}{512}        & 512    & 256   & 512  \\
Sequence   length     & \multicolumn{1}{c}{2000}       & \multicolumn{1}{c}{4000}       & 8000  & 1024 & 1024   \\
Dropout rate          & \multicolumn{5}{c}{0.1} \\
Att. dropout   rate   & \multicolumn{5}{c}{0.1} \\
Pooling mode          & \multicolumn{5}{c}{mean}                                                      \\
Batch size            & \multicolumn{1}{c}{32}         & \multicolumn{1}{c}{32}         & 16    & 256      & 128     \\
Learning rate         & \multicolumn{1}{c}{0.0001}     & \multicolumn{1}{c}{0.0001}     & 0.0002    & 0.0001    & 0.0002     \\
Warmup steps          & \multicolumn{1}{c}{1000}       & \multicolumn{1}{c}{80}         & 800    & 175  & 312      \\
Learning rate   decay & \multicolumn{5}{c}{linear}                                                    \\
Weight decay          & \multicolumn{5}{c}{0}   \\
Evaluation   freq.    & \multicolumn{1}{c}{500}        & \multicolumn{1}{c}{500}        & 1000    & 50  & 500     \\
Num. epochs           & \multicolumn{5}{c}{50k} \\
Num. init   steps     & \multicolumn{1}{c}{1k}         & \multicolumn{1}{c}{3K}         & 3k    & 0 & 3.5k     \\
Num. eval   steps     & \multicolumn{1}{c}{62}         & \multicolumn{1}{c}{200}        & 300   & 20 & 156      \\
Patience              & \multicolumn{5}{c}{10}  \\
Num. features         & \multicolumn{5}{c}{\{64, 128, 256, 512\}}                                                      \\
\bottomrule
\end{tabular}
}
\end{table}

\begin{table*}[ht!]
    \centering
    \caption{Detailed experimental results for all variants on the LRA benchmark over five seeds. We report mean accuracy, training time, and peak memory for feature dimensions ($d$) of 64, 128, 256, and 512. L: ListOps, T: Text, R: Retrieval, I: Image, P: Pathfinder.}
    \label{table:lra_1}
    \resizebox{0.9\linewidth}{!}{%
    \begin{tabular}{l c *{18}{c}}
    \toprule
    & & \multicolumn{6}{c}{Accuracy (\%) $\uparrow$} & \multicolumn{6}{c}{Time (hour) $\downarrow$} & \multicolumn{6}{c}{Memory (GB) $\downarrow$} \\
    \cmidrule(lr){3-8} \cmidrule(lr){9-14} \cmidrule(lr){15-20}
    Model Variant & $d$ & {L} & {T} & {R} & {I} & {P} & {$\mu$} & {L} & {T} & {R} & {I} & {P} & {$\mu$} & {L} & {T} & {R} & {I} & {P} & {$\mu$} \\
    \midrule
    \multirow{4}{*}{PosRF-FastFoodL} & 64 & 31.50 (6.32) & 64.61 (0.29) & 77.13 (0.99) & 38.28 (0.43) & 66.38 (0.71) & 55.58 & 1.04 & 2.08 & 2.17 & 3.95 & 2.02 & 2.25 & 0.77 & 1.54 & 1.50 & 3.01 & 1.54 & 1.67 \\
& 128 & 29.54 (5.46) & 64.61 (0.29) & 77.10 (1.02) & 38.28 (0.43) & 66.38 (0.71) & 55.18 & 1.02 & 2.00 & 2.01 & 3.88 & 1.98 & 2.18 & 0.77 & 1.54 & 1.50 & 3.01 & 1.54 & 1.67 \\
& 256 & 29.54 (5.46) & 64.61 (0.29) & 76.94 (0.77) & 38.28 (0.43) & 66.17 (0.29) & 55.11 & 1.01 & 1.98 & 2.02 & 3.88 & 1.98 & 2.17 & 0.77 & 1.54 & 1.50 & 3.01 & 1.54 & 1.67 \\
& 512 & 31.50 (6.32) & 64.61 (0.29) & 77.13 (0.99) & 38.28 (0.43) & 66.38 (0.71) & 55.58 & 1.02 & 1.99 & 2.02 & 3.89 & 1.98 & 2.18 & 0.77 & 1.54 & 1.50 & 3.01 & 1.54 & 1.67 \\
\cmidrule(l){1-20}
\multirow{4}{*}{PosRF-ORF} & 64 & 37.51 (0.37) & 60.46 (0.34) & 80.68 (0.47) & 34.41 (1.11) & 66.93 (0.59) & 56.00 & 0.48 & 0.80 & 0.81 & 1.52 & 0.80 & 0.88 & 0.76 & 1.51 & 1.42 & 2.99 & 1.50 & 1.64 \\
& 128 & 34.50 (6.49) & 61.10 (1.76) & 80.53 (0.33) & 33.72 (1.03) & 65.52 (4.89) & 55.07 & 0.56 & 1.05 & 1.06 & 2.02 & 1.06 & 1.15 & 1.14 & 2.26 & 2.05 & 4.49 & 2.26 & 2.44 \\
& 256 & 37.31 (0.51) & 62.19 (1.82) & 80.73 (0.59) & 32.82 (0.42) & 64.22 (4.62) & 55.46 & 0.78 & 1.54 & 1.55 & 3.06 & 1.55 & 1.70 & 1.89 & 3.76 & 3.31 & 7.50 & 3.76 & 4.04 \\
& 512 & 37.13 (0.15) & 61.34 (1.23) & 80.30 (0.27) & 33.79 (1.11) & 67.35 (0.31) & 55.98 & 1.22 & 2.55 & 2.49 & 5.07 & 2.53 & 2.77 & 3.40 & 6.77 & 5.82 & 13.52 & 6.77 & 7.26 \\
\cmidrule(l){1-20}
\multirow{4}{*}{PosRF-SORF} & 64 & 25.26 (8.01) & 63.50 (0.63) & 67.91 (0.39) & 28.45 (0.81) & 49.88 (0.13) & 47.00 & 0.48 & 0.80 & 0.81 & 1.52 & 0.80 & 0.89 & 0.76 & 1.51 & 1.42 & 2.99 & 1.50 & 1.64 \\
& 128 & 21.91 (5.98) & 62.06 (1.36) & 66.33 (0.41) & 28.70 (3.87) & 52.37 (5.24) & 46.27 & 0.55 & 1.06 & 1.06 & 2.00 & 1.05 & 1.14 & 1.14 & 2.26 & 2.05 & 4.49 & 2.26 & 2.44 \\
& 256 & 20.63 (3.02) & 61.10 (1.11) & 65.81 (0.78) & 26.52 (2.40) & 50.12 (0.36) & 44.84 & 0.78 & 1.54 & 1.56 & 3.05 & 1.55 & 1.69 & 1.89 & 3.76 & 3.30 & 7.50 & 3.76 & 4.04 \\
& 512 & 20.21 (2.70) & 61.48 (0.40) & 65.12 (1.11) & 25.77 (0.64) & 50.05 (0.28) & 44.53 & 1.22 & 2.56 & 2.50 & 5.07 & 2.54 & 2.78 & 3.40 & 6.77 & 5.82 & 13.52 & 6.77 & 7.26 \\
\cmidrule(l){1-20}
\multirow{4}{*}{PosRF-QMC} & 64 & 37.02 (0.17) & 61.25 (1.67) & 80.71 (0.58) & 34.42 (1.36) & 67.17 (0.35) & 56.11 & 0.48 & 0.80 & 0.81 & 1.52 & 0.81 & 0.89 & 0.76 & 1.51 & 1.42 & 2.99 & 1.50 & 1.64 \\
& 128 & 37.05 (0.16) & 60.93 (0.59) & 80.67 (0.29) & 34.03 (0.94) & 67.25 (0.44) & 55.99 & 0.55 & 1.05 & 1.05 & 1.99 & 1.05 & 1.14 & 1.14 & 2.26 & 2.05 & 4.49 & 2.26 & 2.44 \\
& 256 & 37.22 (0.13) & 61.53 (1.43) & 80.67 (0.32) & 34.47 (0.71) & 62.93 (5.13) & 55.36 & 0.78 & 1.54 & 1.56 & 3.04 & 1.55 & 1.69 & 1.89 & 3.76 & 3.30 & 7.50 & 3.76 & 4.04 \\
& 512 & 37.06 (0.63) & 61.92 (1.91) & 80.66 (0.24) & 34.03 (0.55) & 66.85 (0.70) & 56.10 & 1.22 & 2.56 & 2.50 & 5.07 & 2.53 & 2.78 & 3.40 & 6.77 & 5.82 & 13.52 & 6.77 & 7.26 \\
\cmidrule(l){1-20}
\multirow{4}{*}{PosRF-MM} & 64 & 37.06 (0.10) & 60.35 (1.02) & 80.83 (0.18) & 33.55 (1.07) & 64.76 (4.63) & 55.31 & 0.48 & 0.80 & 0.81 & 1.51 & 0.80 & 0.88 & 0.76 & 1.51 & 1.42 & 2.99 & 1.50 & 1.64 \\
& 128 & 37.13 (0.27) & 62.88 (1.17) & 80.58 (0.44) & 33.82 (0.51) & 67.11 (0.45) & 56.30 & 0.56 & 1.05 & 1.06 & 2.02 & 1.05 & 1.15 & 1.14 & 2.26 & 2.05 & 4.49 & 2.26 & 2.44 \\
& 256 & 37.14 (0.16) & 62.66 (1.48) & 80.10 (0.42) & 33.71 (1.41) & 64.43 (5.05) & 55.61 & 0.77 & 1.54 & 1.54 & 3.03 & 1.55 & 1.69 & 1.89 & 3.76 & 3.30 & 7.50 & 3.76 & 4.04 \\
& 512 & 34.31 (6.23) & 62.78 (1.49) & 80.32 (0.29) & 33.14 (1.20) & 64.97 (4.38) & 55.10 & 1.22 & 2.55 & 2.50 & 5.09 & 2.54 & 2.78 & 3.40 & 6.77 & 5.82 & 13.52 & 6.77 & 7.26 \\
\cmidrule(l){1-20}
\multirow{4}{*}{PosRF-SGQ} & 64 & 34.40 (6.21) & 63.04 (0.70) & 78.11 (0.46) & 37.26 (0.63) & 57.72 (8.55) & 54.10 & 0.48 & 0.80 & 0.81 & 1.52 & 0.80 & 0.88 & 0.76 & 1.51 & 1.42 & 2.99 & 1.50 & 1.64 \\
& 128 & 28.81 (7.77) & 62.30 (0.47) & 78.31 (0.39) & 37.96 (0.76) & 59.71 (7.87) & 53.42 & 0.56 & 1.06 & 1.05 & 2.01 & 1.06 & 1.15 & 1.14 & 2.26 & 2.05 & 4.49 & 2.26 & 2.44 \\
& 256 & 31.58 (7.71) & 62.26 (0.36) & 78.28 (0.27) & 37.90 (0.58) & 57.46 (5.25) & 53.50 & 0.78 & 1.54 & 1.55 & 3.03 & 1.56 & 1.69 & 1.89 & 3.76 & 3.30 & 7.50 & 3.76 & 4.04 \\
& 512 & 31.52 (7.60) & 62.57 (0.64) & 78.02 (0.33) & 37.42 (0.51) & 53.87 (3.06) & 52.68 & 1.22 & 2.55 & 2.49 & 5.07 & 2.54 & 2.78 & 3.40 & 6.77 & 5.82 & 13.52 & 6.77 & 7.26 \\
\midrule
\multirow{4}{*}{OPRF-FastFoodL} & 64 & 37.92 (0.39) & 64.32 (0.61) & 77.86 (1.37) & 38.18 (0.71) & 66.76 (0.39) & 57.01 & 1.06 & 2.06 & 2.11 & 4.02 & 2.05 & 2.26 & 0.84 & 1.68 & 1.64 & 3.26 & 1.68 & 1.82 \\
& 128 & 37.73 (0.28) & 64.32 (0.61) & 78.65 (1.94) & 38.41 (0.66) & 66.76 (0.39) & 57.17 & 1.07 & 2.07 & 2.11 & 4.02 & 2.05 & 2.26 & 0.84 & 1.68 & 1.64 & 3.26 & 1.68 & 1.82 \\
& 256 & 37.90 (0.36) & 64.32 (0.61) & 77.32 (0.22) & 38.41 (0.66) & 66.60 (0.15) & 56.91 & 1.06 & 2.06 & 2.11 & 4.01 & 2.05 & 2.26 & 0.84 & 1.68 & 1.64 & 3.26 & 1.67 & 1.82 \\
& 512 & 37.92 (0.39) & 64.32 (0.61) & 77.86 (1.37) & 38.41 (0.66) & 66.76 (0.39) & 57.05 & 1.06 & 2.06 & 2.12 & 4.05 & 2.07 & 2.27 & 0.84 & 1.68 & 1.64 & 3.26 & 1.68 & 1.82 \\
\cmidrule(l){1-20}
\multirow{4}{*}{\textbf{OPRF-ORF}} & 64 & 38.03 (0.41) & 60.29 (0.58) & 80.81 (0.35) & 34.43 (0.49) & 66.82 (0.74) & 56.08 & 0.55 & 0.92 & 0.93 & 1.76 & 0.93 & 1.02 & 0.86 & 1.70 & 1.65 & 3.37 & 1.70 & 1.86 \\
& 128 & 38.05 (0.78) & 60.18 (0.49) & 81.19 (0.23) & 33.70 (0.57) & 67.24 (0.88) & 56.07 & 0.68 & 1.26 & 1.24 & 2.47 & 1.29 & 1.39 & 1.33 & 2.64 & 2.50 & 5.25 & 2.64 & 2.87 \\
& 256 & 37.67 (0.26) & 60.95 (1.38) & 80.76 (0.50) & 33.29 (0.50) & 64.66 (4.82) & 55.47 & 0.98 & 1.92 & 1.86 & 4.21 & 2.11 & 2.22 & 2.27 & 4.52 & 4.19 & 9.01 & 4.52 & 4.90 \\
& 512 & 37.68 (0.27) & 60.07 (0.49) & 80.61 (0.20) & 34.01 (1.34) & 66.92 (0.26) & 55.86 & 1.63 & 3.34 & 3.11 & 7.72 & 3.55 & 3.87 & 4.15 & 8.28 & 7.58 & 16.53 & 8.28 & 8.96 \\
\cmidrule(l){1-20}
\multirow{4}{*}{OPRF-SORF} & 64 & 27.60 (3.10) & 65.22 (0.18) & 70.73 (1.65) & 35.49 (2.67) & 50.41 (0.50) & 49.89 & 0.55 & 0.93 & 0.94 & 1.77 & 0.93 & 1.02 & 0.86 & 1.70 & 1.65 & 3.37 & 1.70 & 1.86 \\
& 128 & 29.59 (3.77) & 64.81 (0.22) & 77.12 (0.78) & 37.42 (0.47) & 63.98 (5.38) & 54.58 & 0.68 & 1.26 & 1.24 & 2.42 & 1.28 & 1.38 & 1.33 & 2.64 & 2.50 & 5.25 & 2.64 & 2.87 \\
& 256 & 28.33 (5.15) & 64.72 (0.31) & 77.59 (0.95) & 37.34 (1.16) & 63.13 (3.80) & 54.22 & 0.98 & 1.92 & 1.86 & 4.23 & 2.11 & 2.22 & 2.27 & 4.52 & 4.19 & 9.01 & 4.52 & 4.90 \\
& 512 & 25.64 (1.89) & 64.82 (0.27) & 76.97 (0.68) & 37.89 (0.70) & 65.99 (0.50) & 54.26 & 1.63 & 3.34 & 3.10 & 7.73 & 3.54 & 3.87 & 4.15 & 8.28 & 7.58 & 16.53 & 8.28 & 8.96 \\
\cmidrule(l){1-20}
\multirow{4}{*}{OPRF-QMC} & 64 & 38.07 (0.69) & 60.52 (0.62) & 80.45 (0.69) & 34.33 (1.02) & 67.04 (0.18) & 56.08 & 0.55 & 0.92 & 0.94 & 1.77 & 0.94 & 1.03 & 0.86 & 1.70 & 1.65 & 3.37 & 1.70 & 1.86 \\
& 128 & 37.86 (0.40) & 60.87 (1.85) & 80.44 (0.20) & 33.87 (0.69) & 66.81 (0.68) & 55.97 & 0.68 & 1.26 & 1.24 & 2.43 & 1.28 & 1.38 & 1.33 & 2.64 & 2.50 & 5.25 & 2.64 & 2.87 \\
& 256 & 37.55 (0.06) & 60.16 (0.65) & 80.45 (0.22) & 34.35 (0.40) & 62.75 (5.33) & 55.05 & 0.98 & 1.93 & 1.87 & 4.24 & 2.12 & 2.23 & 2.27 & 4.52 & 4.19 & 9.01 & 4.52 & 4.90 \\
& 512 & 37.83 (0.58) & 59.84 (0.87) & 80.24 (0.83) & 33.80 (0.56) & 66.57 (0.90) & 55.66 & 1.64 & 3.34 & 3.11 & 7.74 & 3.53 & 3.87 & 4.15 & 8.28 & 7.58 & 16.53 & 8.28 & 8.96 \\
\cmidrule(l){1-20}
\multirow{4}{*}{OPRF-MM} & 64 & 37.90 (0.66) & 60.29 (0.61) & 80.97 (0.31) & 33.61 (0.62) & 66.08 (2.11) & 55.77 & 0.55 & 0.92 & 0.93 & 1.76 & 0.93 & 1.02 & 0.86 & 1.70 & 1.65 & 3.37 & 1.70 & 1.86 \\
& 128 & 38.31 (0.38) & 60.01 (1.01) & 81.03 (0.31) & 33.93 (0.85) & 65.17 (5.17) & 55.69 & 0.68 & 1.26 & 1.29 & 2.47 & 1.28 & 1.40 & 1.33 & 2.64 & 2.50 & 5.25 & 2.64 & 2.87 \\
& 256 & 37.59 (0.47) & 60.27 (0.67) & 80.94 (0.46) & 33.47 (0.87) & 63.15 (5.61) & 55.09 & 0.98 & 1.92 & 1.87 & 4.23 & 2.10 & 2.22 & 2.27 & 4.52 & 4.19 & 9.01 & 4.52 & 4.90 \\
& 512 & 37.73 (0.25) & 60.39 (2.04) & 80.82 (0.28) & 33.49 (0.76) & 66.43 (1.27) & 55.77 & 1.63 & 3.33 & 3.09 & 7.71 & 3.54 & 3.86 & 4.15 & 8.28 & 7.58 & 16.53 & 8.28 & 8.96 \\
\cmidrule(l){1-20}
\multirow{4}{*}{OPRF-SGQ} & 64 & 37.66 (0.52) & 61.25 (0.50) & 78.68 (0.83) & 36.64 (1.33) & 59.77 (6.20) & 54.80 & 0.55 & 0.92 & 0.93 & 1.76 & 0.93 & 1.02 & 0.86 & 1.70 & 1.65 & 3.37 & 1.70 & 1.86 \\
& 128 & 37.08 (0.56) & 61.09 (0.77) & 79.39 (0.94) & 36.68 (0.28) & 67.53 (2.79) & 56.35 & 0.68 & 1.27 & 1.25 & 2.46 & 1.29 & 1.39 & 1.33 & 2.64 & 2.50 & 5.25 & 2.64 & 2.87 \\
& 256 & 37.16 (0.28) & 61.36 (0.36) & 79.24 (0.11) & 36.34 (1.11) & 65.11 (5.13) & 55.84 & 0.98 & 1.92 & 1.87 & 4.22 & 2.11 & 2.22 & 2.27 & 4.52 & 4.19 & 9.01 & 4.52 & 4.90 \\
& 512 & 37.20 (0.51) & 61.37 (0.25) & 79.27 (0.37) & 36.91 (1.24) & 64.75 (4.87) & 55.90 & 1.63 & 3.34 & 3.10 & 7.72 & 3.53 & 3.86 & 4.15 & 8.28 & 7.58 & 16.53 & 8.28 & 8.96 \\
\midrule
\multirow{4}{*}{SADERF-FastFoodL} & 64 & 29.72 (7.27) & 64.71 (0.45) & 75.92 (1.17) & 38.38 (0.82) & 65.97 (0.88) & 54.94 & 1.09 & 2.12 & 2.16 & 4.05 & 2.05 & 2.30 & 0.90 & 1.80 & 1.75 & 3.52 & 1.80 & 1.95 \\
& 128 & 29.72 (7.27) & 64.71 (0.45) & 77.50 (0.96) & 38.38 (0.82) & 66.47 (1.47) & 55.35 & 1.09 & 2.09 & 2.18 & 4.05 & 2.06 & 2.29 & 0.90 & 1.80 & 1.76 & 3.52 & 1.80 & 1.96 \\
& 256 & 29.72 (7.27) & 64.71 (0.45) & 76.25 (0.73) & 38.05 (1.18) & 65.97 (0.88) & 54.94 & 1.07 & 2.07 & 2.16 & 4.04 & 2.05 & 2.28 & 0.90 & 1.80 & 1.75 & 3.52 & 1.80 & 1.95 \\
& 512 & 29.72 (7.27) & 64.71 (0.45) & 76.64 (1.59) & 38.38 (0.82) & 65.97 (0.88) & 55.08 & 1.06 & 2.07 & 2.16 & 4.06 & 2.06 & 2.29 & 0.90 & 1.80 & 1.76 & 3.52 & 1.80 & 1.95 \\
\cmidrule(l){1-20}
\multirow{4}{*}{\textbf{SADERF-ORF}} & 64 & 37.19 (0.61) & 60.33 (2.13) & 80.83 (0.25) & 33.68 (1.23) & 66.96 (0.81) & 55.80 & 0.56 & 0.92 & 0.98 & 1.76 & 0.92 & 1.03 & 0.93 & 1.85 & 1.78 & 3.68 & 1.85 & 2.02 \\
& 128 & 37.41 (0.45) & 59.92 (0.90) & 81.05 (0.18) & 33.06 (1.02) & 67.11 (0.56) & 55.71 & 0.69 & 1.25 & 1.28 & 2.50 & 1.28 & 1.40 & 1.40 & 2.79 & 2.63 & 5.56 & 2.79 & 3.04 \\
& 256 & 33.98 (6.41) & 60.05 (0.78) & 80.96 (0.17) & 33.27 (0.89) & 62.91 (5.16) & 54.23 & 0.96 & 1.92 & 1.93 & 4.26 & 2.10 & 2.23 & 2.34 & 4.67 & 4.32 & 9.32 & 4.67 & 5.07 \\
& 512 & 34.45 (6.19) & 61.68 (2.08) & 80.62 (0.16) & 33.31 (1.60) & 64.53 (4.48) & 54.92 & 1.59 & 3.34 & 3.13 & 7.75 & 3.53 & 3.87 & 4.23 & 8.43 & 7.71 & 16.84 & 8.43 & 9.13 \\
\cmidrule(l){1-20}
\multirow{4}{*}{SADERF-SORF} & 64 & 26.42 (4.53) & 65.22 (0.20) & 68.74 (1.05) & 33.39 (1.21) & 50.31 (0.49) & 48.82 & 0.56 & 0.92 & 0.98 & 1.76 & 0.92 & 1.03 & 0.93 & 1.85 & 1.78 & 3.68 & 1.85 & 2.02 \\
& 128 & 34.31 (2.50) & 64.82 (0.26) & 75.24 (1.12) & 36.76 (1.22) & 60.98 (4.90) & 54.42 & 0.68 & 1.25 & 1.28 & 2.47 & 1.26 & 1.39 & 1.40 & 2.79 & 2.63 & 5.56 & 2.79 & 3.04 \\
& 256 & 32.36 (6.74) & 64.92 (0.18) & 76.65 (1.04) & 36.73 (0.78) & 58.76 (3.42) & 53.88 & 0.96 & 1.91 & 1.91 & 4.25 & 2.10 & 2.23 & 2.34 & 4.67 & 4.32 & 9.32 & 4.67 & 5.07 \\
& 512 & 31.60 (2.32) & 64.85 (0.16) & 76.99 (0.49) & 37.23 (1.08) & 61.54 (4.79) & 54.44 & 1.60 & 3.35 & 3.14 & 7.74 & 3.52 & 3.87 & 4.23 & 8.43 & 7.71 & 16.84 & 8.43 & 9.13 \\
\cmidrule(l){1-20}
\multirow{4}{*}{SADERF-QMC} & 64 & 37.35 (0.40) & 60.76 (1.74) & 80.93 (0.20) & 34.55 (1.23) & 66.60 (1.77) & 56.04 & 0.56 & 0.91 & 0.99 & 1.78 & 0.93 & 1.04 & 0.93 & 1.85 & 1.78 & 3.68 & 1.85 & 2.02 \\
& 128 & 37.18 (0.47) & 60.61 (2.06) & 80.67 (0.44) & 34.55 (0.63) & 67.32 (0.40) & 56.07 & 0.68 & 1.24 & 1.28 & 2.47 & 1.26 & 1.39 & 1.40 & 2.79 & 2.63 & 5.56 & 2.79 & 3.04 \\
& 256 & 37.42 (0.45) & 60.01 (0.63) & 80.94 (0.31) & 34.50 (0.64) & 64.89 (4.75) & 55.55 & 0.96 & 1.92 & 1.91 & 4.25 & 2.09 & 2.23 & 2.34 & 4.67 & 4.32 & 9.32 & 4.67 & 5.07 \\
& 512 & 36.95 (0.46) & 59.56 (1.23) & 80.36 (0.76) & 34.21 (1.28) & 61.78 (5.53) & 54.57 & 1.60 & 3.34 & 3.13 & 7.75 & 3.56 & 3.88 & 4.23 & 8.43 & 7.71 & 16.84 & 8.43 & 9.13 \\
\cmidrule(l){1-20}
\multirow{4}{*}{SADERF-MM} & 64 & 34.65 (5.94) & 60.48 (3.13) & 80.76 (0.29) & 32.92 (0.95) & 64.85 (4.74) & 54.73 & 0.56 & 0.92 & 0.98 & 1.76 & 0.92 & 1.03 & 0.93 & 1.85 & 1.78 & 3.68 & 1.85 & 2.02 \\
& 128 & 37.17 (0.27) & 60.81 (1.82) & 81.05 (0.20) & 34.03 (1.04) & 65.31 (4.85) & 55.67 & 0.69 & 1.25 & 1.28 & 2.49 & 1.28 & 1.40 & 1.40 & 2.79 & 2.63 & 5.56 & 2.79 & 3.04 \\
& 256 & 37.15 (0.64) & 60.69 (1.64) & 80.83 (0.19) & 33.06 (0.82) & 62.91 (5.97) & 54.93 & 0.95 & 1.91 & 1.91 & 4.24 & 2.08 & 2.22 & 2.34 & 4.67 & 4.32 & 9.32 & 4.67 & 5.07 \\
& 512 & 37.69 (0.74) & 61.38 (2.25) & 80.81 (0.26) & 33.63 (0.84) & 64.97 (4.43) & 55.70 & 1.59 & 3.33 & 3.13 & 7.73 & 3.53 & 3.86 & 4.23 & 8.43 & 7.71 & 16.84 & 8.43 & 9.13 \\
\cmidrule(l){1-20}
\multirow{4}{*}{SADERF-SGQ} & 64 & 34.19 (6.27) & 62.69 (0.75) & 77.96 (0.68) & 37.81 (0.43) & 59.31 (6.87) & 54.39 & 0.56 & 0.92 & 0.98 & 1.77 & 0.92 & 1.03 & 0.93 & 1.85 & 1.78 & 3.68 & 1.85 & 2.02 \\
& 128 & 37.22 (0.28) & 62.86 (0.96) & 78.51 (0.68) & 37.82 (0.59) & 61.27 (5.38) & 55.54 & 0.68 & 1.25 & 1.28 & 2.50 & 1.27 & 1.39 & 1.40 & 2.79 & 2.63 & 5.56 & 2.79 & 3.04 \\
& 256 & 37.18 (0.31) & 63.66 (0.22) & 77.94 (0.44) & 37.62 (0.55) & 62.61 (3.12) & 55.80 & 0.97 & 1.94 & 1.93 & 4.28 & 2.11 & 2.24 & 2.34 & 4.67 & 4.32 & 9.32 & 4.67 & 5.07 \\
& 512 & 37.08 (0.10) & 62.92 (0.79) & 78.40 (0.80) & 37.89 (0.30) & 65.79 (0.84) & 56.42 & 1.60 & 3.35 & 3.14 & 7.75 & 3.53 & 3.88 & 4.23 & 8.43 & 7.71 & 16.84 & 8.43 & 9.13 \\\bottomrule
    \end{tabular}%
    }
    \end{table*}

\begin{table*}[h]
\caption{
    Summary statistics across component functions ($f$) and weight matrices ($W$) for 64, 128, 256, and 512 feature settings.
    The original four tables have been consolidated for clarity and compactness.
    We report mean accuracy (test set \%), mean training time (hours), and mean peak memory consumption (GB).
    Tasks are abbreviated as $L$: ListOps, $T$: Text, $R$: Retrieval, $I$: Image, $P$: Pathfinder, and $\mu$ for the average.
}
\label{table:summary_stats}
\centering
\resizebox{\linewidth}{!}{
\begin{tabular}{@{}c l r *{18}{r}@{}}
\toprule
\multirow{2}{*}{Type} & \multirow{2}{*}{Method} & \multirow{2}{*}{Features} & \multicolumn{6}{c}{Accuracy (\%) $\uparrow$} & \multicolumn{6}{c}{Time (hour) $\downarrow$} & \multicolumn{6}{c}{Memory (GB) $\downarrow$} \\
\cmidrule(lr){4-9} \cmidrule(lr){10-15} \cmidrule(lr){16-21}
& & & {L} & {T} & {R} & {I} & {P} & {$\mu$} & {L} & {T} & {R} & {I} & {P} & {$\mu$} & {L} & {T} & {R} & {I} & {P} & {$\mu$} \\
\midrule
\multirow{12}{*}{$f$} & \multirow{4}{*}{PosRF} & 64 & 33.79 & 62.20 & 77.56 & 34.39 & 62.14 & 54.02 & 0.57 & 1.02 & 1.04 & 1.93 & 1.01 & 1.11 & 0.76 & 1.51 & 1.43 & 2.99 & 1.51 & 1.64 \\
& & 128 & 31.49 & 62.32 & 77.25 & 34.42 & 63.06 & 53.71 & 0.63 & 1.21 & 1.21 & 2.32 & 1.21 & 1.32 & 1.08 & 2.14 & 1.96 & 4.25 & 2.14 & 2.31 \\
& & 256 & 32.24 & 62.39 & 77.09 & 33.95 & 60.89 & 53.31 & 0.82 & 1.61 & 1.63 & 3.18 & 1.62 & 1.77 & 1.70 & 3.39 & 3.00 & 6.75 & 3.39 & 3.65 \\
& & 512 & 31.95 & 62.45 & 76.89 & 33.74 & 61.58 & 53.16 & 1.19 & 2.46 & 2.41 & 4.88 & 2.44 & 2.68 & 2.96 & 5.90 & 5.07 & 11.77 & 5.90 & 6.33 \\
\cmidrule(lr){2-21}
& \multirow{4}{*}{OPRF} & 64 & 36.20 & 61.98 & 78.25 & 35.45 & 62.81 & 54.94 & 0.63 & 1.11 & 1.13 & 2.14 & 1.12 & 1.23 & 0.86 & 1.70 & 1.65 & 3.35 & 1.69 & 1.85 \\
& & 128 & 36.44 & 61.88 & 79.64 & 35.67 & 66.25 & 55.97 & 0.74 & 1.40 & 1.40 & 2.71 & 1.41 & 1.53 & 1.25 & 2.48 & 2.36 & 4.92 & 2.48 & 2.70 \\
& & 256 & 36.03 & 61.97 & 79.38 & 35.53 & 64.23 & 55.43 & 0.99 & 1.94 & 1.91 & 4.19 & 2.10 & 2.23 & 2.03 & 4.05 & 3.77 & 8.05 & 4.04 & 4.39 \\
& & 512 & 35.67 & 61.80 & 79.30 & 35.75 & 66.24 & 55.75 & 1.54 & 3.12 & 2.94 & 7.11 & 3.29 & 3.60 & 3.60 & 7.18 & 6.59 & 14.32 & 7.18 & 7.77 \\
\cmidrule(lr){2-21}
& \multirow{4}{*}{SADERF} & 64 & 33.25 & 62.36 & 77.53 & 35.12 & 62.34 & 54.12 & 0.65 & 1.12 & 1.18 & 2.15 & 1.11 & 1.24 & 0.93 & 1.84 & 1.78 & 3.66 & 1.84 & 2.01 \\
& & 128 & 35.50 & 62.29 & 79.01 & 35.77 & 64.74 & 55.46 & 0.75 & 1.39 & 1.43 & 2.75 & 1.40 & 1.54 & 1.32 & 2.63 & 2.48 & 5.22 & 2.63 & 2.86 \\
& & 256 & 34.63 & 62.34 & 78.93 & 35.54 & 63.01 & 54.89 & 0.98 & 1.94 & 1.96 & 4.22 & 2.09 & 2.24 & 2.10 & 4.19 & 3.89 & 8.36 & 4.19 & 4.55 \\
& & 512 & 34.58 & 62.52 & 78.97 & 35.86 & 64.10 & 55.33 & 1.51 & 3.13 & 2.97 & 7.11 & 3.29 & 3.58 & 3.67 & 7.32 & 6.72 & 14.54 & 7.33 & 7.87 \\
\midrule
\multirow{24}{*}{$W$} & \multirow{4}{*}{FastFoodL} & 64 & 33.05 & 64.55 & 76.97 & 38.28 & 66.37 & 55.84 & 1.06 & 2.09 & 2.15 & 4.01 & 2.04 & 2.27 & 0.84 & 1.67 & 1.63 & 3.26 & 1.67 & 1.82 \\
& & 128 & 32.33 & 64.55 & 77.75 & 38.35 & 66.54 & 55.90 & 1.06 & 2.05 & 2.10 & 3.99 & 2.03 & 2.25 & 0.84 & 1.67 & 1.63 & 3.26 & 1.67 & 1.82 \\
& & 256 & 32.38 & 64.55 & 76.83 & 38.25 & 66.25 & 55.65 & 1.05 & 2.03 & 2.10 & 3.98 & 2.03 & 2.24 & 0.84 & 1.67 & 1.63 & 3.26 & 1.67 & 1.82 \\
& & 512 & 33.05 & 64.55 & 77.21 & 38.35 & 66.37 & 55.91 & 1.05 & 2.04 & 2.10 & 4.00 & 2.04 & 2.25 & 0.84 & 1.67 & 1.63 & 3.26 & 1.67 & 1.82 \\
\cmidrule(lr){2-21}
& \multirow{4}{*}{MM} & 64 & 36.54 & 60.37 & 80.85 & 33.36 & 65.23 & 55.27 & 0.53 & 0.88 & 0.91 & 1.68 & 0.88 & 0.97 & 0.85 & 1.69 & 1.62 & 3.35 & 1.68 & 1.84 \\
& & 128 & 37.53 & 61.23 & 80.89 & 33.93 & 65.86 & 55.89 & 0.64 & 1.19 & 1.21 & 2.33 & 1.21 & 1.32 & 1.29 & 2.56 & 2.39 & 5.10 & 2.56 & 2.78 \\
& & 256 & 37.29 & 61.21 & 80.62 & 33.41 & 63.50 & 55.21 & 0.90 & 1.79 & 1.77 & 3.83 & 1.91 & 2.04 & 2.17 & 4.32 & 3.94 & 8.61 & 4.32 & 4.67 \\
& & 512 & 36.58 & 61.52 & 80.65 & 33.42 & 65.46 & 55.52 & 1.48 & 3.07 & 2.91 & 6.85 & 3.20 & 3.50 & 3.92 & 7.83 & 7.03 & 15.63 & 7.83 & 8.45 \\
\cmidrule(lr){2-21}
& \multirow{4}{*}{ORF} & 64 & 37.58 & 60.36 & 80.78 & 34.17 & 66.90 & 55.96 & 0.53 & 0.88 & 0.91 & 1.68 & 0.89 & 0.98 & 0.85 & 1.69 & 1.62 & 3.35 & 1.68 & 1.84 \\
& & 128 & 36.65 & 60.40 & 80.92 & 33.49 & 66.62 & 55.62 & 0.64 & 1.19 & 1.19 & 2.33 & 1.21 & 1.31 & 1.29 & 2.56 & 2.39 & 5.10 & 2.56 & 2.78 \\
& & 256 & 36.32 & 61.06 & 80.82 & 33.13 & 63.93 & 55.05 & 0.91 & 1.80 & 1.78 & 3.84 & 1.92 & 2.05 & 2.17 & 4.32 & 3.94 & 8.61 & 4.32 & 4.67 \\
& & 512 & 36.42 & 61.03 & 80.51 & 33.73 & 66.27 & 55.89 & 1.48 & 3.08 & 2.91 & 6.78 & 3.20 & 3.45 & 3.92 & 7.83 & 7.03 & 15.54 & 7.83 & 8.34 \\
\cmidrule(lr){2-21}
& \multirow{4}{*}{QMC} & 64 & 37.48 & 60.84 & 80.70 & 34.43 & 66.93 & 56.08 & 0.53 & 0.88 & 0.91 & 1.69 & 0.89 & 0.98 & 0.85 & 1.69 & 1.62 & 3.35 & 1.68 & 1.84 \\
& & 128 & 37.36 & 60.80 & 80.59 & 34.15 & 67.13 & 56.01 & 0.64 & 1.18 & 1.19 & 2.30 & 1.20 & 1.30 & 1.29 & 2.56 & 2.39 & 5.10 & 2.56 & 2.78 \\
& & 256 & 37.40 & 60.57 & 80.68 & 34.44 & 63.52 & 55.32 & 0.91 & 1.79 & 1.78 & 3.84 & 1.92 & 2.05 & 2.17 & 4.32 & 3.94 & 8.61 & 4.32 & 4.67 \\
& & 512 & 37.28 & 60.44 & 80.42 & 34.01 & 65.07 & 55.44 & 1.49 & 3.08 & 2.91 & 6.86 & 3.21 & 3.51 & 3.92 & 7.83 & 7.03 & 15.63 & 7.83 & 8.45 \\
\cmidrule(lr){2-21}
& \multirow{4}{*}{SGQ} & 64 & 35.42 & 62.33 & 78.25 & 37.23 & 58.93 & 54.43 & 0.53 & 0.88 & 0.91 & 1.68 & 0.88 & 0.98 & 0.85 & 1.69 & 1.62 & 3.35 & 1.68 & 1.84 \\
& & 128 & 34.37 & 62.08 & 78.74 & 37.49 & 62.84 & 55.10 & 0.64 & 1.19 & 1.19 & 2.32 & 1.21 & 1.31 & 1.29 & 2.56 & 2.39 & 5.10 & 2.56 & 2.78 \\
& & 256 & 35.31 & 62.43 & 78.49 & 37.29 & 61.72 & 55.05 & 0.91 & 1.80 & 1.79 & 3.84 & 1.93 & 2.05 & 2.17 & 4.32 & 3.94 & 8.61 & 4.32 & 4.67 \\
& & 512 & 35.27 & 62.29 & 78.60 & 37.41 & 61.47 & 54.69 & 1.48 & 3.08 & 2.94 & 6.85 & 3.20 & 3.52 & 3.92 & 7.83 & 7.12 & 15.63 & 7.83 & 8.48 \\
\cmidrule(lr){2-21}
& \multirow{4}{*}{SORF} & 64 & 26.43 & 64.65 & 69.13 & 32.44 & 50.20 & 48.57 & 0.53 & 0.88 & 0.91 & 1.68 & 0.89 & 0.98 & 0.85 & 1.69 & 1.62 & 3.35 & 1.68 & 1.84 \\
& & 128 & 28.60 & 63.90 & 72.90 & 34.29 & 59.11 & 51.76 & 0.64 & 1.19 & 1.19 & 2.30 & 1.20 & 1.30 & 1.29 & 2.56 & 2.39 & 5.10 & 2.56 & 2.78 \\
& & 256 & 27.10 & 63.58 & 73.35 & 33.53 & 57.34 & 50.98 & 0.91 & 1.79 & 1.78 & 3.84 & 1.92 & 2.05 & 2.17 & 4.32 & 3.94 & 8.61 & 4.32 & 4.67 \\
& & 512 & 25.82 & 63.71 & 73.03 & 33.63 & 59.19 & 51.08 & 1.49 & 3.08 & 2.91 & 6.85 & 3.20 & 3.51 & 3.92 & 7.83 & 7.03 & 15.63 & 7.83 & 8.45 \\\bottomrule
\end{tabular}
}
\end{table*}

\begin{figure}[ht]
    \centering
    \includegraphics[width=\linewidth]{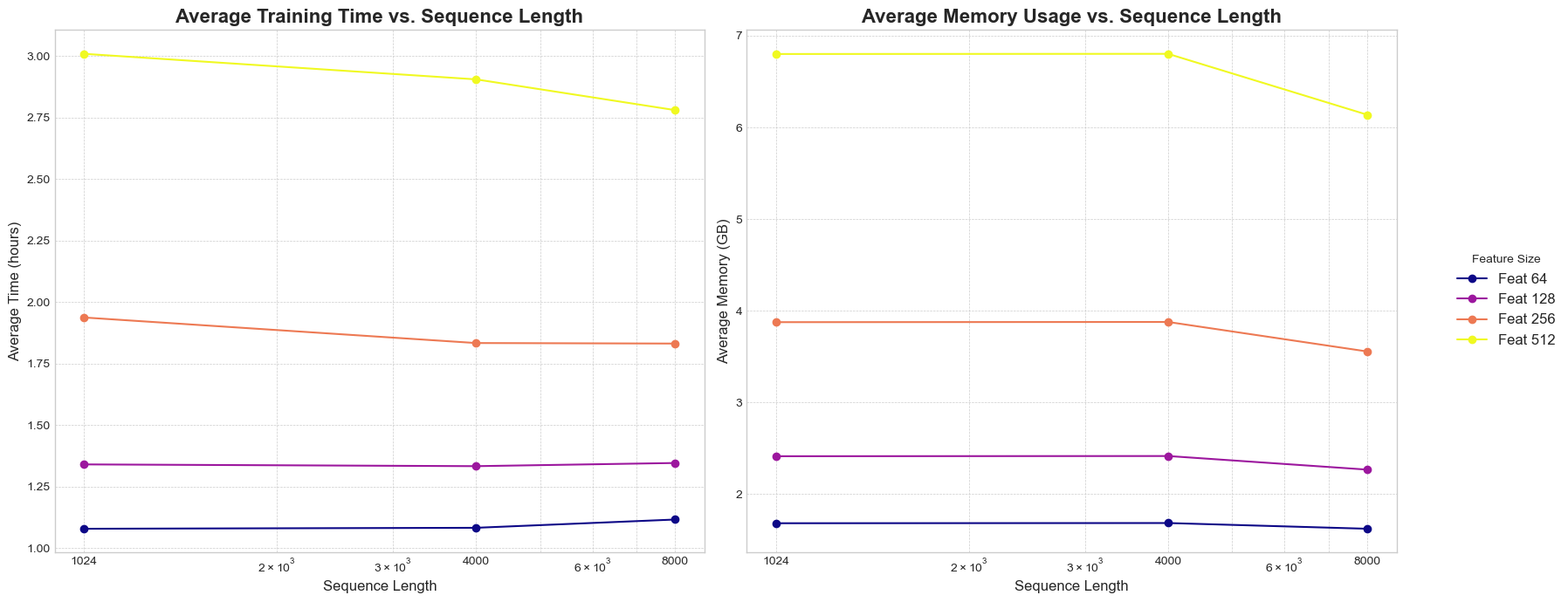}
    \caption{Efficiency of Training Time and Memory Usage w.r.t. Sequence Length for Different Number of Features. Only Text, Retrieval and Pathfinder data are used due to having the same number of labels (2) compared to ListOps and Image (10).}
    \label{fig:efficiency}
\end{figure}

\end{document}